\documentclass[preprint,12pt]{elsarticle}

\usepackage{times}
\usepackage{epsfig}
\usepackage{graphicx}
\usepackage{amsmath}
\usepackage{amssymb}
\usepackage{epstopdf}
\usepackage{tabularx}
\usepackage{multirow}
\usepackage{enumerate}
\usepackage{wrapfig}
\usepackage{array}
\usepackage{hyperref}
\usepackage{algorithm2e}
\usepackage{xcolor,colortbl}
\usepackage{hhline}

\usepackage{color,ulem}

\newcommand{\red}{\textcolor[rgb]{0,0.0,0.0}}

\definecolor{LightCyan}{rgb}{1,0.9,0.9}

\DeclareMathOperator*{\argmin}{arg\,min}
\DeclareMathOperator*{\argmax}{arg\,max}
\newcolumntype{C}[1]{>{\centering\let\newline\\\arraybackslash\hspace{0pt}}m{#1}}

\journal{Computer Vision and Image Understanding}

\begin{document}

\begin{frontmatter}

\title{You-Do, I-Learn: Egocentric Unsupervised \\Discovery of Objects and their Modes of Interaction\\ Towards Video-Based Guidance}

\author{Dima Damen}
\author{Teesid Leelasawassuk}
\author{Walterio Mayol-Cuevas}

\address{
 Computer Science Department\\
 University of Bristol, Bristol, UK
}

\begin{abstract}
This paper presents an unsupervised approach towards automatically extracting video-based guidance on object usage, from egocentric video and wearable gaze tracking, collected from multiple users while performing tasks.
The approach i)~discovers task relevant objects, ii) builds a model for each, iii)~distinguishes different ways in which each discovered object has been used and iv)~discovers the dependencies between object interactions.
The work investigates using appearance, position, motion and attention, and presents results using each and a combination of relevant features. 
Moreover, an online scalable approach is presented and is compared to offline results. 
The paper proposes a method for selecting a suitable video guide to be displayed to a novice user indicating how to use an object, purely triggered by the user's gaze.
The potential assistive mode can also recommend an object to be used next based on the learnt sequence of object interactions.
The approach was tested on a variety of daily tasks such as initialising a printer, preparing a coffee and setting up a gym machine.
\end{abstract}


\end{frontmatter}

\section{Introduction}
\label{sec:intro}
\begin{figure}[t]
\centering
\includegraphics[width=1.0\columnwidth]{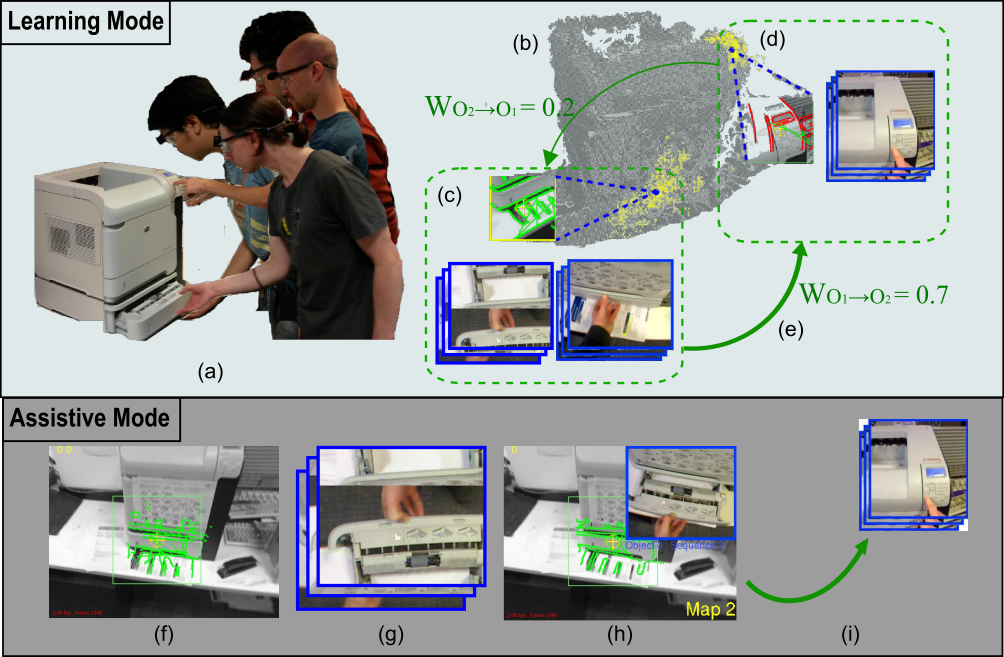}
\caption{Given egocentric videos from multiple users (a), a map of the environment (b) and feature clustering are used to discover distinct task-relevant objects (TROs); e.g. paper drawer and keypad for the task of operating a printer (c,d). For each discovered TRO, a model is built to incorporate possible locations, appearance and usage, along with a probabilistic graph of object interactions (e). In a potential \textit{assistive mode}, when a TRO is recognised triggered by gaze (f), a usage snippet can be chosen~(g) and can be displayed to the user to provide guidance on how to use the object~(h) along with the most likely object to be used next~(i).}
\label{fig:overview}
\end{figure}

Increasingly, commercial interest in wearable devices, including cameras and head-mounted displays in miniature and in fully wearable form (e.g. Google's Glass, Microsoft's HoloLens, Sony's SmartEyeglass) invited research into cognitive systems that take advantage of these platforms.
Footage from wearable cameras has fuelled Internet-based video sharing sites. 
Interestingly, among the most sought after videos are \textit{how to do} guides, accessed by people wishing to carry out tasks, from cooking to assembling furniture.

Assistance in task performance (e.g. assembly, repair) using augmented
reality or video guidance has been promised for a while. One
of the key limitations to realise such systems is the time consuming and evidently limiting task of authoring the content by e.g.
manually segmenting and annotating videos or creating
three-dimensional models that represent meaningful guidance (e.g.~\cite{Petersen12},\cite{Bleser13}).
Approaches that can discover object interactions from video input and provide guidance without the need for manual intervention would enable a wider adoption of assistive wearable systems.

Figure~\ref{fig:overview} shows an overview of the proposed \textit{You-Do, I-Learn} approach, both the learning and the assistive modes.
This work attempts, to \textbf{fully unsupervised}, discover objects and their usage from multiple users in a common environment (Fig.~\ref{fig:overview}a), then proposes a complete automatic solution for object-based guidance for discovered objects.
\textit{Note that the suggested assistive mode can potentially be implemented on a wearable device, but this is out of the scope of this work.}

In proposing the \textit{You-Do, I-Learn} approach, we particularly focus on an egocentric view of the world, taking advantage of wearable technology, as it offers a unique perspective on object-level interactions.
As opposed to discovering all objects in the environment, we focus on discovering task relevant objects.
A {\textbf {Task Relevant Object (TRO)}} is an object, or part of an object, with which a person interacts during task performance. For example, a person operating a printer may interact with the paper drawer (Fig.~\ref{fig:overview}c) and/or the keypad (Fig.~\ref{fig:overview}d) while operating it. A system that aims to discover TROs would attempt to discover these objects/parts as opposed to the full machine or all of its parts.
For each discovered object, we build a location model, an appearance model
and collect usage snippets on how different users interacted with the same discovered object.
A {\textbf{usage snippet}} is an automatically extracted video sequence, to reflect how an object has been used.
Several usage snippets can be extracted for the same TRO as the object is used multiple times by the same or different users.

The various models of TROs can be used to provide assistive guidance. The location model guides the user to where an object can be found. The appearance model is used to recognise the object when visible in the wearable device's field of view. The collected usage snippets can be used for video-based guidance. 
To achieve video-based guidance on object usage, we also introduce the term {\textbf {Modes of Interaction (MOI)}} to refer to the different ways in which TROs are used. Say, a cup can be lifted, washed, or poured into.
All these are different MOIs associated to the cup. When harvesting usage snippets for the same object from multiple operators, common MOIs can be discovered.
In introducing MOIs, we distinguish between object-based guidance and task-based guidance.
This is because the same object can be used in many tasks, while the ways in which one object can be interacted with are usually limited to a finite set of possible interactions.
Triggered simply by gaze (Fig.~\ref{fig:overview}f), the user is advised on how a TRO object can be used based on the object's current state (Fig.~\ref{fig:overview}h), as well as advise the user on the most-likely object to be used next (Fig.~\ref{fig:overview}i).

Section~\ref{sec:review} presents an overview of previous attempts towards object discovery, in general and for egocentric video in particular, as well as attempts towards video-based guidance.
The learning and assistive modes are presented in Sections~\ref{sec:learning} and~\ref{sec:assistive} respectively.
A varied dataset from coffee preparation to operating a gym machine is presented, alongside results in Section~\ref{sec:results}.
\red{Section~\ref{sec:models} discusses building approximate three-dimensional models for discovered objects, as a by-product of the approach. 
These can be used for virtual reality guidance, but this is left for future work.} 
The paper concludes with future directions.

\section{Video-Based Object Discovery and Guidance - a Review}
\label{sec:review}

\begin{figure}[h!]
  \begin{center}
    \includegraphics[width=0.47\textwidth]{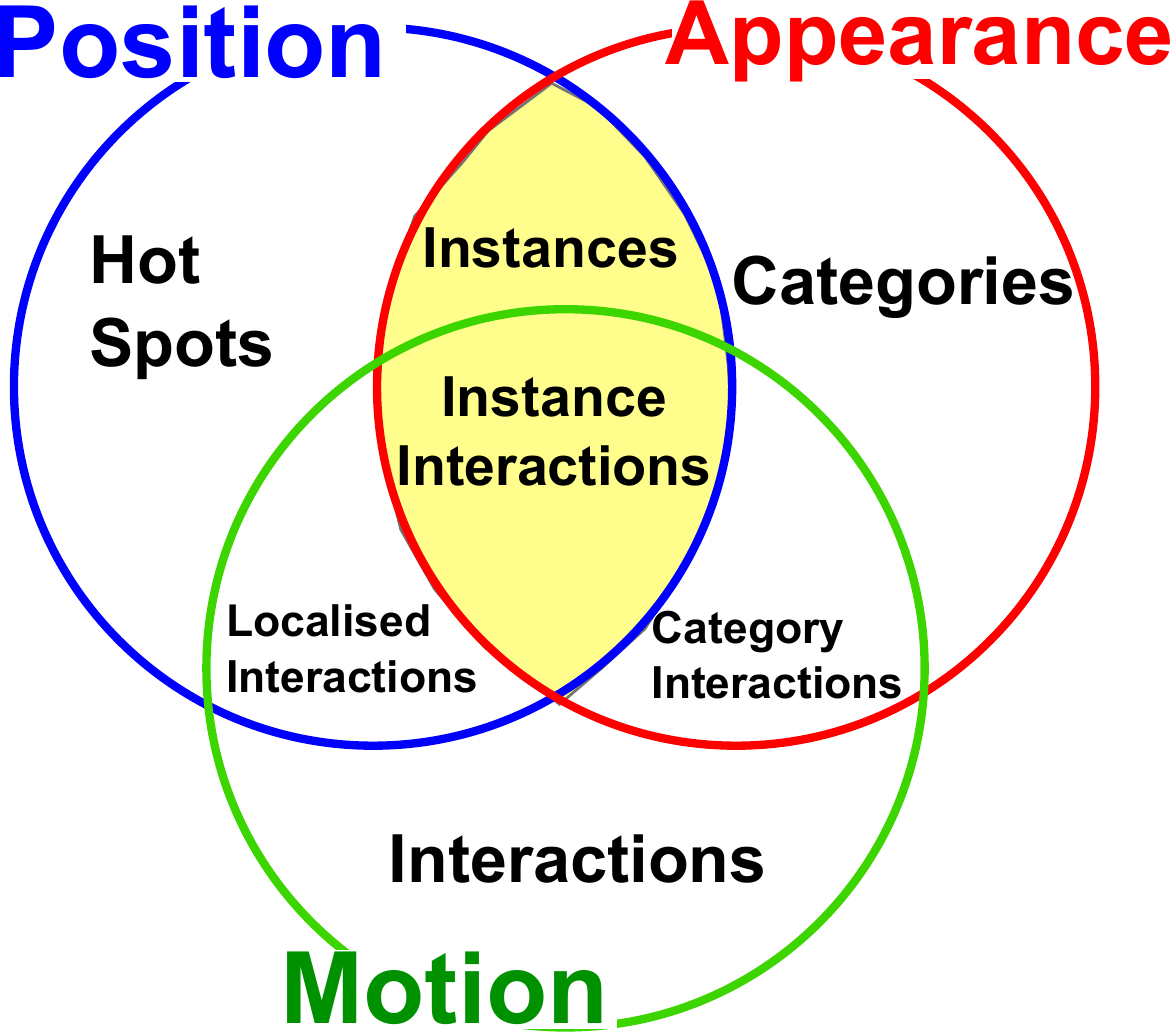}
  \end{center}
\caption{Using appearance, position, motion and combinations for object discovery}
\label{fig:plot}
\end{figure}
Object discovery refers to grouping feature descriptors into meaningful clusters that correspond to entities worth discovering.
We attempt to differentiate the various ways in which entities can be discovered from video input; appearance, position and motion.
Figure~\ref{fig:plot} envisages what can be discovered if each, or a combination, of these information is used in the grouping.
The \textbf{position}, relative to an environment, can be grouped into hot-spots.
A \textit{hot spot} is a position at which object interaction takes place, and could correspond to objects that remain fixed relative to the environment.
\textbf{Appearance} similarity is a strong cue to discover visual categories, i.e. objects that share similar appearance.
When combining appearance with position, instances of objects can be discovered.
In video, \textbf{motion} presents a third cue that could be employed solely to discover ego actions and interactions. When combined with instances, different manners of interaction with objects can be discovered.

\red{The number of works attempting to use egocentric vision for tasks ranging from attention estimation to activity recognition has increased exponentially in the last decade.
The reader can refer to early works \cite{Mayol09, Mayol04} or more recent surveys~\cite{Betancourt15, Nguyen16} for a review of datasets, methods and attempted problems.}
In this review, we focus on \textit{unsupervised} approaches to object discovery (Section~\ref{sec:related-discovery}), task-relevant object discovery from egocentric video (Section~\ref{sec:related-trodiscovery}) as well as approaches that aim to link discovery to guidance (Section~\ref{sec:related-similarApproaches}).
The works presented here differ from the frequent attempts to recognise objects in egocentric video from supervised training (e.g. ~\cite{Gonzalez14, Pirsiavash12, Ren09}) which are interesting in their own right.

\subsection{Unsupervised Object Discovery from Static Images}
\label{sec:related-discovery}
Appearance similarity, along with the objects' positions, has been deployed in various works to discover objects from a set of images or 3D scenes in an unsupervised manner.

\noindent\textbf{Appearance:}\hspace{4pt}  Appearance, of object and context, is often used to discover \textit{categories} (e.g. \cite{Kim08b, Russell06, Tuytelaars10}) or \textit{instances} (e.g. \cite{Karpathy13, Kang15, Kang11, Sanders02}) of an object based on visual similarity.
When attempting to discover categories, most works assume a collection of images where \textit{common} features correspond to one category, with or without spatial consistency (e.g.~\cite{Kim08b, Russell06}).
These approaches are only semi-supervised as collecting images belonging to a category is needed.
In~\cite{Tuytelaars10}, Tuytelaars \textit{et al} review and compare recent works in category-based discovery from a dataset of images.

Unsupervised instance discovery, similar to our work, has also been attempted from a set of images and 3D scenes~\cite{Karpathy13, Kang15, Kang11}.
In~\cite{Karpathy13}, 3D object segments are discovered from a dataset of indoor {RGB-D} scenes.
Several measures assist segmentation: compactness, symmetry, local convexity, global convexity, smoothness and recurrence in multiple scenes.
Object segment hypotheses are accordingly ranked to come up with a final set of discovered objects.
In~\cite{Kang15}, a data-driven objectness measure is proposed where a segment is compared to a database of general object segments.
In~\cite{Kang11}, colour, texture and shape-based features are used to construct a network of finely-segmented regions.
Segments are then iteratively grouped and refined until the algorithm converges to discovered objects.
While a very interesting approach with promising results, \cite{Kang11} assumes that objects of daily living are moveable.
A computer screen, for example, needs to be moved to a different background to enable its discovery.
Many objects of daily living such as a coffee machine or an electric socket remain fixed to their surroundings.
The approaches in \cite{Kang15, Kang11} also assume the dataset contains at least one instance of an object of interest per image.
When using video as input, a significant number of frames might not contain TROs as the user roams around an environment.
Moreover, these approaches are processed offline after the dataset is collected.

\noindent\textbf{Position:}\hspace{4pt}  
Position information has been used solely in~\cite{Herbst11} to discover objects, by aligning two point clouds and identifying changes in location that correspond to objects that have been placed or removed.
In~\cite{Mason12}, the disappearance of features in a position is used as a cue to discover objects from RGB-D images collected using a mobile platform.
This assumption is also considered by an earlier work~\cite{Sanders02} which uses multiple cameras to build a depth map.

\noindent\textbf{Appearance and Position:}\hspace{4pt} 
Combining position and appearance cues has been used to enable discover objects~\cite{Somanath09, Collet13}.
In \cite{Somanath09}, sensor tracking enables constructing a model of objects placed on a planar surface. Segments are combined from multiple scenes using appearance matching of interest points.
A database of models is used to refine the reconstruction of discovered objects.
In~\cite{Collet13}, RGB-D images collected from a robot in a common environment are first separated into discrete locations (rooms, in their case), then appearance and depth data are clustered to extract instances.
The approach assumes that all objects are placed on a planar surface (e.g. table-top) and employs a prior on the object's shape and size.

\subsection{Unsupervised Discovery of Task-Relevant Objects from Egocentric Video}
\label{sec:related-trodiscovery}
Egocentric video introduces two new sources of information to object discovery; motion and attention. 
Motion is the result of the wearer's self-motion or objects in an environment. 
Egocentric video shows a vantage viewpoint to objects the person attends to.
This section reviews works in egocentric video analysis towards, or related to, unsupervised discovery of objects.

\noindent\textbf{Motion:}\hspace{4pt}  In egocentric video, motion descriptors have been proposed for action recognition, either full-body actions~\cite{Kitani11} or object interactions~\cite{Sundaram09, Sundaram12}.
In~\cite{Yan14}, unsupervised discovery of object interactions is attempted, by clustering unlabelled video snippets representing actions.
The problem is formulated as a linear program and a solver is used to find the optimal clustering, using the earth mover's distance measure.
Though unsupervised in nature, the approach assumes the number of tasks (i.e. the number of clusters) is known, based on the knowledge that the people in the dataset perform a pre-specified set of tasks.
The approach compares K-means, Kernel K-means as well as convex and semi-nonnegative matrix factorisation.

\noindent\textbf{Appearance, Motion and Attention:}\hspace{4pt} 
As opposed to discovering all objects, several works focus on discovering objects with which the person interacts, whether towards object discovery or action recognition.

In a recent work, Bolanos proposed a semi-supervised approach to discover objects from wearable sensors~\cite{Bolanos15}. The video is uniformly sampled into a sparse set of images at $\frac{1}{60}$fps.
Given partial labelling, objectness measures along with Convolutional Neural Networks (CNN) are used as features with iterative clustering.
Clustering is evaluated, using silhouette coefficients, to decide on the discovered objects.

In~\cite{Fathi13}, an interaction is identified by the change in appearance of the object before and after the action is performed.
Foreground segmentation is used, followed by extracting the hand-held object regions.
Unsupervised clustering of object segments enables modelling the change in the object's appearance.

In~\cite{Lee12}, objects of `importance' are segmented from egocentric video sequences using appearance and motion features.
Segmentation is based on the similarity to `segments of importance' from a manually labelled training set, collected via crowd sourcing.

Accordingly, common approaches to discovering TROs in egocentric vision include i) segmenting the area connected the user's hand~\cite{Fathi11, Fathi13, Lee12}, ii) extracting foreground regions through frame stabilisation or scene planarity assumptions~\cite{Ren10, Sundaram12} or iii) detecting `object-like' regions~\cite{Lu13, Bolanos15}.
The first two approaches are only able to segment objects while being manipulated, during which objects could be heavily occluded by the hand.
In the second approach, fixed objects like a sink tap or a coffee machine, which can be quite crucial to a task, are ignored.
In the third approach, `object-like' regions can focus on salient rather than used objects. 

Very few systems exploit the high quality and predictive nature of eye gaze fixation.
Its anticipatory nature allows estimating which object will be used next~\cite{Land99, Land06}.
Gaze has been successfully used to assist action and activity recognition~\cite{Fathi12, Li13, Ogaki12, Matsuo15} and supervised object recognition~\cite{Sun09, DeBeugher12}.

\subsection{Unsupervised Video-Based Guidance}
\label{sec:related-similarApproaches}
Unsupervised extraction of video snippets from a continuous egocentric video has mostly targeted video summarisation~\cite{Lu13, Lee12}.
The earliest example we could trace of segmenting video snippets for guidance is the work of Kang and Ikeuchi in 1994~\cite{Kang94} that uses stereo visual data and other sensors, and focuses on tracking hand motion. The extracted snippets are used for guidance of robotic arms during grasping.
Similarly, the work of Mayol and Murray in 2005~\cite{Mayol05} automatically detects keyframes of interactions with objects from a shoulder-mounted camera, towards event-based summarisation.

For human user assistance, Hashimoto et al.~\cite{Hashimoto11} proposed view sharing of video from wearable cameras to guide novice users.
Their work does not require unsupervised segmenatation of video guides but focuses on live sharing of egocentric views.
In~\cite{Goto10}, instructional videos are projected onto an AR display for task guidance.
While manually edited instruction video clips are employed, the system paces the instruction clips to match the status of the performed task.
In~\cite{Petersen12}, automatic extraction of snippets is performed using novelty detection. Video clips are extracted based on the distance between consecutive frames.
The work also discusses overlaying the segmentation videos onto the scene in real time.

\vspace{12pt}

Up to our knowledge, this manuscript presents the first attempt to close the gap between object discovery and video-based guidance in a fully unsupervised way.
The manuscript builds on our previous works towards offline~\cite{Damen14} and online~\cite{Damen14b} discovery of task relevant objects, with additional novel contributions:
\begin{itemize} 
\item A generalised formulation for the problem of discovering task-relevant objects from a sequence of egocentric images.

\item An improved online discovery algorithm compared to the one proposed in~\cite{Damen14b} with superior performance. The new algorithm uses a Gaussian Mixture Model (GMM) to represent each discovered object as opposed to a threshold over Euclidean distance from~\cite{Damen14b}.

\item Previously unpublished comparison of the online approach to offline discovery of TROs.

\item Building a graph of object interactions, which can provide guidance on which object to use next in a sequence of object interactions.

\item A detailed explanation of how subject annotations can be used to generate ground-truth of task-relevant objects and their usages.
\end{itemize}
\noindent Moreover, the paper provides further details on both the offline and the online algorithms.
The approach is explained next; first the learning mode (Sec.~\ref{sec:learning}) then the assistive mode (Sec.~\ref{sec:assistive}).

\section{You-Do, I-Learn: Learning Mode}
\label{sec:learning}
During learning, Task-Relevant Objects (TRO) need to be discovered (Sec.~\ref{sec:discovery}) and a model to be built for each object (Sec.~\ref{sec:model}).
For each discovered TRO, usage snippets are automatically collected showing multiple people interacting with the same object.
These usage snippets can be analysed to discover the various Modes of Interaction (MOI) in an unsupervised manner (Sec.~\ref{sec:moi}).
Sequences of object interactions can also be discovered, highlighting strong temporal dependencies (Sec.~\ref{sec:graph}).

\subsection{Discovering Task Relevant Objects (TRO)}
\label{sec:discovery}
We first present a formulation for the problem of discovering task-relevant objects from egocentric video. Given a sequence of egocentric images $\{I_1, .., I_T\}$ collected from multiple operators around a common environment, TRO discovery is the process of finding $K$ TROs, $\{O_k; 1 \le k \le K \}$, where the number of objects $K$ is not known \textit{a priori}. 
Assume $\Omega(I_t)$ is a part of the image $I_t$ (e.g. a segmentation or a bounding box within $I_t$), each discovered TRO $O_k$ is a set of image parts from the sequence.
\begin{equation}
O_k = \{ \Omega(I_t); 1 \le t \le T \}
\end{equation}

\begin{figure}[t]
\centering
\includegraphics[width=1.0\columnwidth]{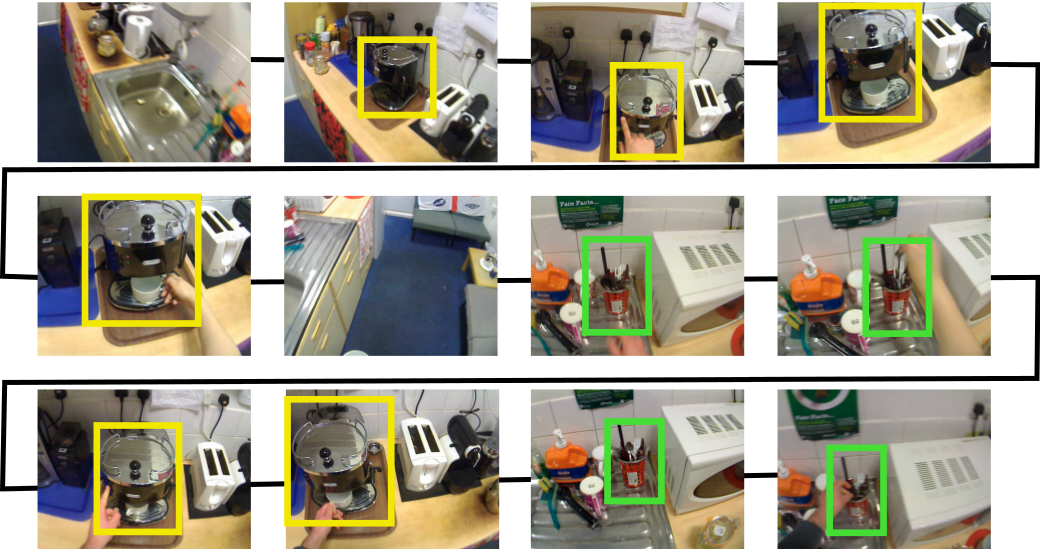}
\caption{Given a sequence of images from egocentric views, the objective is to group image parts into TROs, based on the assumption that one image part at most is task-relevant in each image. 
In this example, two TROs are shown.}
\label{fig:TRO_ex}
\end{figure}
In this formulation, we make the assumption that at most one task-relevant image part is present within each image $I_t$.
\red{The notion of `at most' handles cases when the person is not actively interacting with any object in the environment.
During interactions, only one image part is task-relevant.}
The person could be interacting with multiple objects, for example placing one object on top of another, yet the attention is believed to shift between these objects~\cite{Johansson01}.
This assumption simplifies the discovery of TROs without much loss in generality.
Accordingly, the sets of image parts representing discovered TROs $\{O_k\}$ are believed to be disjoint and form a subset of all images in the sequence.
Figure~\ref{fig:TRO_ex} shows a visual representation of TRO discovery formulation.

Given the above formulation, we next propose two approaches to TRO discovery, one is offline assuming all sequences from multiple users are collected prior to the discovery.
The second approach is online, and thus is scalable to multiple users and different TROs.
For image parts $\Omega(I_t)$, we only report results on using bounding boxes.
The proposed online and offline approaches are applicable to segmentations, but this is left for future work.
We compare two techniques to suggest a bounding box in an image that can contain a TRO.
The first $\Omega_c(I_t)$ crops the image around the centre. Given a glass-mounted camera, it is expected to have the object of interest at the centre of the frame during interactions.
We compare this approach to gaze fixation $\Omega_g(I_t)$ where the image is cropped around a known gaze fixation.
Using a wearable gaze tracker, we filter saccades using the velocity-based approach from~\cite{Salvucci2000}, where the average angular velocity over a sliding temporal window is considered a saccade if it is greater than $100^\circ/sec$, and is thus discarded.
\red{Note that we do not use image-based saliency to find the image part $\Omega(I_t)$. There are potentially many visually salient objects in an environment that are not interacted with.
Our interest is to discover only those objects, whether visually salient or otherwise, with which one interacts.}
 
To describe image parts, we use position and appearance features as well as their combination following Fig.~\ref{fig:plot}:
\begin{itemize}
\item \textbf{Position:}\hspace{4pt} The Image $I_t$ is positioned relative to the scene using sparse Simultaneous Localisation and Mapping (SLAM)~\cite{Klein2007}. A triangular tessellation of tracked interest points is built, similar to~\cite{Takemura2010}.
Given the 6D pose of the scene camera, a 3D ray connects the centre of the image part $\Omega(I_t)$ to a point in the scene.
Using the tessellation, the 3D position of the intersection point is calculated using linear interpolation.
\textit{Using position features solely enables discovering static TROs. Moveable objects, observed in different locations will be discovered as separate TROs.}

\item \textbf{Appearance:}\hspace{4pt} To represent appearance, Histogram of Oriented Gradients (HOG)~\cite{Dalal05} is calculated over sub-windows within the image patch $\Omega(I_t)$.
In offline TRO discovery, Bag of Words (BoW) representation is used for appearance information.
In online TRO discovery, HOG features are used as appearance features directly. 
This is because BoWs require either a representative prior training sequence or an adaptive approach that can merge and introduce new words.
Generalisation of an adaptive BoW approach to the variety of locations and tasks we report in the experiments section would not be trivial.
\textit{Using appearance features solely enables discovering moveable objects, as clusters combine observations of a similar appearance. Static objects though could be separated into multiple clusters if varying viewpoints are observed.}

\item \textbf{Combining Position and Appearance:}\hspace{4pt} When combining position and appearance features, the normalised affinity matrices are summed with equal weighting in offline TRO discovery. The features are simply combined for online TRO discovery. \textit{By combining position and appearance features, static objects can be discovered and moveable objects can be combined using appearance feature similarity.}

\end{itemize}

We also compare to results that accumulate features over a sliding window $w$ centred around each image $(\Omega(I_{t-\frac{w-1}{2}}),..,\Omega(I_t),..,\Omega(I_{t+\frac{w-1}{2}}))$. In the experiments section, we test features that use position, appearance and their combination, over a sliding window and the two image part methods.
We use the term $f_t$ next to refer to the feature vector representing an image part where,
\begin{equation}
f_t = (\digamma(\Omega(I_{t-\frac{w-1}{2}})),..,\digamma(\Omega(I_t)),..,\digamma(\Omega(I_{t+\frac{w-1}{2}}))
\end{equation}
and $\digamma(\Omega(I_t))$ is the feature descriptor for the image part $\Omega(I_t)$.

\subsubsection{Offline TRO Discovery}
\label{sec:offline}
Offline TRO discovery refers to the attempt to discover all TROs after the dataset is fully collected.
The sequencing of images is thus discarded and a data point $x_i = f_t$ refers to the descriptor of an image part in the dataset.
We compare k-means clustering to spectral clustering from Ng~\textit{et al.}~\cite{Ng02}.
These approaches were compared in~\cite{Tuytelaars10} for a known number of object categories.

Unsupervised discovery, like other grouping problems, suffers from the dilemma of model selection (i.e. the optimal number of groups).
Most previous approaches assume the number of groupings is known \textit{a priori}~\cite{Kim08b, Tuytelaars10} to avoid the complexity.
We propose estimating the optimal number of clusters $\hat{K}$ using the Davies-Bouldin (DB) index~\cite{Davies79}.
For an object $O_k$ with $n_k$ data points $\{x_i; i = 1..n_k\}$ assigned to this cluster, and $\mu_k$ is the mean of these data points, the intra-cluster distance $S_k$ can be measured as (Euclidean distance used):
\begin{equation}
S_k = \sqrt{\frac{1}{n_k} \sum_{i=1}^{n_k}{{||x_i - \mu_k||}_2}}
\end{equation}
The inter-cluster distances between two objects $O_k$ and $O_j$ is measured as $M_{kj} = {||\mu_k - \mu_j||}_2$.
The cluster similarity measure $R_{kj} = \frac{S_k+S_j}{M_{kj}}$ is used to calculate DB index,
\begin{equation}
V_{DB}(K) = \frac{1}{K}\sum_{k=1}^K{\max\limits_{j \ne k} R_{kj}}
\label{eq:bd_index}
\end{equation}
The optimal number of clusters is calculated to be 
\begin{equation}
\hat{K} = \argmax_{K} V_{DB}(K)
\end{equation}

Recall that some images do not contain a TRO (Fig.~\ref{fig:TRO_ex}), while clustering assigns a label for each data point.
We assign a probability to each cluster being a TRO as the ratio of the number of points in the cluster to the total number of points,
\begin{equation}
p(S_k) = \frac{n_k}{\sum\limits_{j=1}^{\hat{K}}{n_j}}
\end{equation}
Clusters are refined by removing the furthest $\beta\%$ of points in the cluster from the mean $\mu_k$.
The refinement threshold, $\beta$, was set to 75 in all experimental results.

\subsubsection{Online TRO Discovery}
\label{sec:online}
\setlength{\algomargin}{1.5em}
{\fontsize{8}{8} \selectfont
\restylealgo{boxed}
\linesnumbered
\begin{algorithm}[h!]
\footnotesize{
  \SetVline
  \SetCommentSty{textsf}
  \dontprintsemicolon
  \SetKwInOut{Input}{input}
  \SetKwInOut{Output}{output}
  \SetKwFor{ForEach}{foreach}{}
  
  \AlFnt
  \Input{Image parts and feature vectors $\{(\Omega(I_t), f_t)\}; t=1..T$}
  \Output{TROs $\{O_k; 1 \le k \le K\}$ where $O_k = (\{\Omega(I_t)\}, \Phi_k)$ and $\Phi_k = \{(\theta_{ki}, \mu_{ki}, \Sigma_{ki}); i=1..L_k\}$}
  \BlankLine
  \BlankLine                   
                   $K$ = 0\;
                   $candidate$ = 0\;
                   \For {$t = 1..T$} 
                   {
                   find closest cluster $k$: $\min\arg_k \sum_{i=1}^{L_k} {\theta_{ki} ||f_t - \mu_{ki}||_{\Sigma_{ki}}}$\;
          	   \If{$\sum_{i=1}^{L_k} {\theta_{ki} ||f_t - \mu_{ki}||_{\Sigma_{ki}}} \le \epsilon_2$}
			 {
                           $l = \min\arg_l ||f_t - \mu_{kl}||_{\Sigma_{kl}}; 1 \le l \le L_k$\; 
			    Update $\theta_{kl}$, $\mu_{kl} (Eq~\ref{eq:gp3}), \Sigma_{kl} (Eq~\ref{eq:gp4})$\;
                         			 }
			 \Else
			 {
                            \If{$||f_t - f_{t-1}|| < \epsilon_1$}
                            {
                                $candidate$ = $candidate$ + 1\;
                                \If{$candidate \ge \xi$}
                                {
                                           $K = K + 1$\;
                                           $L_K = 1$\;
                                           Calculate $\mu_K$ and $\Sigma_K$\;
                                           
                                }
                               
                            }
                            \Else
                            {
                                $candidate$ = 0 \;
                            }
			 }

\If{$\min_{j \ne k} d_B(O_k, O_j) < \epsilon_3$}
                            {

 					$L_j = L_j + 1$\;
	                                $\mu_{jL_j} = \mu_{k}$\;
          	                        $\Sigma_{jL_j} = \Sigma_{k}$\;
                    	                Calculate mixture components $\theta_j$\;
                                Delete $O_k$ (objects merged)\;
                                K = K - 1\;
                            }

         }}
  \caption{Proposed algorithm for \textit{online} TRO discovery}
  \label{algo:learn}
\end{algorithm}
}

To discover objects in an online manner, image parts are clustered as they are collected and clusters are incrementally updated.
An approach for online TRO discovery should iteratively cluster image parts of the same object as the object is used by multiple operators, whether in the same or a different location.

In proposing an algorithm for online TRO discovery, we rely on the assumption that \textit{consecutive similar image parts $(\Omega(I_{t - \xi + 1}), .., \Omega(I_t))$ indicate an observation of a task-relevant object (TRO)}.
We thus define a TRO $O_k$ as a collection of `at least' $\xi$ \textit{consecutive and similar} image parts.
The notion of similarity relies on the features used.
For example, when $f_t$ is the 3D position of image part $\Omega(I_t)$, then at least $\xi$ spatially-close consecutive image parts are labelled as a TRO.
Alternatively when $f_t$ is the appearance of image part $\Omega(I_t)$, then at least $\xi$ consecutive image parts of similar appearance enable discovering a TRO.

Two consecutive image parts, $\Omega(I_t)$ and $\Omega(I_{t-1})$ belong to the same object if $||f_{t} - f_{t-1}|| < \epsilon_1$, where $\epsilon_1$ is the threshold selected to accept clustering consecutive image parts and $||.||$ is the Euclidean distance (Algo.~\ref{algo:learn} L. 9-14).
The strict consecutive constraint between $t$ and $t-1$ can be relaxed to allow proximity within a sliding window.
The mean and covariance of $O_k$ are updated incrementally as further image parts are located within the threshold $\epsilon_1$.
Equations~\ref{eq:gp3} and~\ref{eq:gp4} show the incremental update for the mean and covariance of a $O_k$.

\begin{eqnarray}
||f_t - f_{t-1}|| < \epsilon_1 &\rightarrow &\mu^k_t = \frac{\mu^k_{t-1}\times (n_t^k-1) + f_{t}}{n_t^k}
\label{eq:gp3}\\
&\rightarrow &\Sigma^k_t = \frac{n_t^k-2}{n_t^k-1} \Sigma^k_{t-1} + \frac{1}{n_t^k}(f_t - \mu_{n_t^k-1}^k)^T(f_t - \mu_{n_t^k-1}^k)
\label{eq:gp4}
\end{eqnarray}
where $\mu_t^k$ is the mean, $\Sigma_t^k$ is the covariance matrix and $n_t^k$ is the number of image parts within $O_k$ at time $t$.

Attention is believed to have moved to another object when $||f_{t} - f_{t-1}|| \ge \epsilon_1$.
At a future point in time $t+\rho$, further image parts $\Omega(I_{t+\rho})$ can belong to the same TRO $O_k$ if it is within $\epsilon_2$ standard deviations from the TRO $k$ using the Mahalanobis distance (Algo.~\ref{algo:learn} L. 4-5).
This clustering method does not pre-define the size of the clusters.
When using position as a feature, it enables both small-sized and large TROs to be discovered.

The algorithm enables combining observations of the same object, in different locations, into the same cluster (Algo.~\ref{algo:learn} L. 17).
When using appearance-based similarity, observations of moveable objects are grouped together. 
Two clusters $(\mu_t^j, \Sigma_t^j)$ and $(\mu_t^k, \Sigma_t^k)$ are merged if the distance measure $d_B$ is below a threshold $\epsilon_3$.
We use Bhattacharyya distance over appearance features for merging clusters.
As multiple observations of a moveable object $O_k$ are not necessarily close in spatial location, a Gaussian Mixture Model (GMM) $\{(\theta_i, \mu_i, \Sigma_i), i=1..L_k\}$ is used to represent the location model where $\theta_i$ is the mixture component of the Gaussian $i$ and $L_k$ is the number of Gaussians in the GMM for object $k$ (Algo.~\ref{algo:learn} L. 18-23).
A new Gaussian is added to the GMM every time an object of similar appearance is found in a new position.

\subsection{Building Models of TROs}
\label{sec:model}
\begin{figure}[h!]
\centering
\includegraphics[width=1.0\columnwidth]{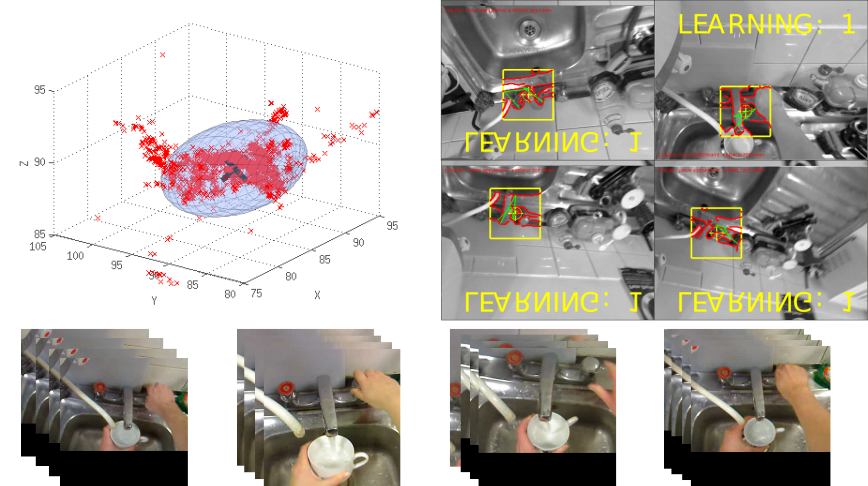}
\caption{\red{For a discovered TRO (tap): multiple users enable modelling the object's position $\Phi_k$ (top-left) learning varying views in the appearance model $A_k$ (top-right)
and gathering different usage snippets $U_k$ that show interactions with the same object (bottom).}}
\label{fig:multi-user}
\end{figure}

Section~\ref{sec:discovery} proposed an offline as well as an online approach for discovering task relevant objects from egocentric video.
For each discovered object $O_k$, \red{we build three models that encapsulate the object's location $\Phi_k$, appearance $A_k$ as well as its usage $U_k$.}
As the models are built from multiple operators with different heights and interaction behaviours, they give a good representation of the object (e.g. Fig.~\ref{fig:multi-user}).
These models can enable a broad range of potential assistance to users. The location model guides the user to where an object can be found. 
The appearance model is used to recognise the object when seen again. The collected usage snippets are used for video-based guidance in Section~\ref{sec:moi}. 
We next detail how the various models can be built for each discovered TRO.

\noindent \textbf{Location Model $\Phi_k$:}\hspace{4pt} The location model represents the position and extent of the object using a Gaussian Mixture Model (GMM) $\Phi_k$; a single Gaussian for a fixed object and a multi-variate Gaussian for moveable objects.
In the offline approach, position features are clustered and the DB-index is used to decide on the number of Gaussians in the GMM. In the online approach, a new Gaussian is introduced for every observation in a new location.
The likelihood of the object's position is then,
\begin{equation}
P(f_t | O_k) = \sum_{l = 1}^{L_k} \theta_{kl} e^{[-\frac{1}{2}(f_t - \mu_{kl})^T \Sigma_{kl}^{-1} (f_t - \mu_{kl})]}
\label{eq:locationLikelihood}
\end{equation}

\noindent \textbf{Appearance Model $\Phi_k$:}\hspace{4pt}For a view-based appearance model, we use the real-time method from~\cite{Damen12} for learning novel views of the object.
This method is particularly helpful for online learning as it is scalable and works in real-time.
It is shape-based and thus is particularly suitable for texture-minimal objects, many of which are present in indoor environments.

\noindent \textbf{Usage Model $U_k$:}\hspace{4pt} The consecutive image parts clustered into one TRO are combined into video snippets indicating how a TRO was used by multiple operators.
Given consecutive image parts $\{\Omega(I_t), \Omega(I_{t+1}), \Omega(I_{t + \rho})\}; \rho \ge \xi$ belonging to the same TRO, a {\textbf {usage snippet $u_i^k$}} is a video clip formed by combining the sequence of image parts.
Notice that when using gaze fixations $\Omega_g$, interpolation is needed when gaze information is missing.
The collection of all usage snippets $U_k = \{u_i^k\}$ shows different ways in which $O_k$ was used or interacted with. These usage snippets are further analysed in Section~\ref{sec:moi} for discovering the various modes of interaction with the same TRO.

\subsection{Finding Modes of Interaction (MOI) for TROs}
\label{sec:moi}
For discovered TROs, we find common MOIs for each TRO by analysing usage snippets, each representing a sample usage.
The collection of all usage snippets $U_k= \{u_i^k\}$ shows different ways in which $O_k$ was used.
Position and appearance information of all frames in $u_i$ (superscript $k$ removed for simplicity) are the same features used for discovering objects.
These are augmented with motion information collected using the Histogram of Optical Flow (HOF) descriptors around 3D Harris points~\cite{Laptev05} to encode the interaction with the object.

We also use a \textit{temporal pyramid} to encode the descriptors.
At each level $l$, the snippet is split into $l$ equally-sized temporal segments, and the descriptor is calculated for each segment.
The temporal pyramid could potentially separate MOIs that differ in their temporal ordering, such as opening and closing.
A one-dimensional representation of the temporal pyramid formulates the descriptor $d(u_i)$.
Clustering then follows (as in~\ref{sec:offline}) to find the MOIs.

Each cluster is represented by the snippet $\hat{u_j}$ closest to the centre of the cluster $\mu_j$ (i.e. mean snippet),
\begin{equation}
\hat{u_j} = \argmin_{u_l \in {MOI}_j} ||d(u_l)-\mu_j||;
\quad
\mu_j = \frac{1}{|{MOI}_j|}\sum_{u_l \in {MOI}_j}{d(u_l)}
\end{equation}
and the confidence in a cluster being a common mode of interaction is represented by the percentage of snippets within that cluster $p({MOI}_j)$,
\begin{equation}
p({MOI}_j) = \frac{|{MOI}_j|}{|U_k|}; \quad \quad p({MOI}_j) \ge \lambda
\label{eq:moi}
\end{equation}
A threshold $\lambda$ can be used to select common MOIs such that $p({MOI}_j) \ge \lambda$.

\subsection{Graphs of Object Interactions}
\label{sec:graph}
Following the discovery of TROs, it is also possible to model, in an unsupervised way, the sequence of object interactions towards modelling tasks or simply discovering strong links between object interactions.
For example, after using the tap, the user is likely to follow that by interacting with the drainer or with a towel.
These strong links between objects can be automatically discovered from sequences of multiple users.
While a more complex interaction model can be targeted, we employ the first-order Markovian assumption.
We model TRO interaction sequences by a graph-based representation. 

For all discovered objects $\{O_k; k=1..K\}$, a \textit{complete} directed graph $G$ is constructed so each TRO is represented by a node and the weight $W_{O_k \rightarrow O_j}$ of the directed edge $O_k \rightarrow O_j$ represents the probability of interacting with object $O_j$ directly after having interacted with object $O_k$.
Note that we loosely define interaction as attending or looking at an object.
\red{The edge weights are initialised with a small value $\alpha$. 
Temporal transitions from one discovered object to another are accounted for, followed by edge-weight normalisation.}

\section {You-Do, I-Learn: Assistive Mode}
\label{sec:assistive}

In the assistive mode, the location models $\{\Phi_k\}$, the appearance models $\{A_k\}$, the usage snippets $\{U_k\}$, the various modes of interaction $\{MOI_k\}$ as well as the graph of object interactions $G_{K \times K}$ are used to provide a recommendation of how an object can be used, as well as what object to use next.

To provide guidance, the object with which the user is attempting to interact should be recognised. In a test image $I_t$, the image part $\Omega(I_t)$ is compared to the discovered TROs.
Upon recognition of a TRO $O_k$, video-based guidance can be provided by showing one of the possible MOIs, that is most relevant to the task or the object status.
The {\it help snippet} $h_t = u_k \in U_k$ is chosen from the possibly many {\it usage snippets} featuring the TRO.
We choose the {\it help snippet} $h_t$ as a usage guide at time $t$ such that the appearance of the first frame in the snippet, is closest to the recognised view.
\begin{equation}
h_t = \arg\min_{u_j} ||A^{1st}(u_j)-A(\Omega(I_t))||
\label{eq:help}
\end{equation}
\noindent where $A^{1st}$ is the appearance of the first frame in the snippet.
If the object changes state, the initial appearance is a good indicator of which usage snippet to show.
An additional advantage is to avoid showing a snippet observing the object from a different viewpoint, so the user can easily map what they see to what they could do. 
A {\it help snippet} is displayed each time a new object is detected, aiming to provide automatic assistance for novice operators.

The graph of object interactions $G_{K \times K}$ can be used to estimate the object to be next manipulated.
The assistive mode would recommend the object to be used next, so that
\begin{equation}
\hat{j} = \arg \max_j p(O_j | O_k) = \arg \max_j W_{O_k \rightarrow O_j}
\end{equation}
\noindent The location model for the recommended object $O_{\hat{j}}$ can be used to suggest where the object is likely to be found (Eq.~\ref{eq:locationLikelihood}), 
\begin{equation}
\hat{l_j} = \arg \max_{f_t} P(f_t | O_{\hat{j}})
\end{equation}
Recommending a help snippet, the object to be used next as well as where that object can be found are based on correctly recognising that the user is attending a TRO. We base the assistive mode on gaze fixations $\Omega_g(I_t)$ and investigate two approaches for recognising the TRO,
\begin{enumerate}
\item \textbf{Using the location models $\{\Phi_k\}$}: a TRO is recognised based on Eq.~\ref{eq:locationLikelihood} so that 
\begin{equation}
k = \arg\max_k p(f_t | O_k) ; \quad \quad p(f_t | O_k) \ge \lambda
\end{equation}

\item \textbf{Using the appearance models $\{A_k\}$}: this assistive mode does not require a map of the environment or tracking of the camera relative to the environment.
Given the image part $\Omega_g(I_t)$, the appearance model is used to recognise the viewed object, from the set of appearance models.
By using the combination of fixed paths and a hierarchical hash table, object recognition is scalable, and can reliably detect objects at frame rate~\cite{Damen12}.
The descriptor is affine-invariant, and the method is tolerant to a level of occlusion but is also view-dependant.
Figure~\ref{fig:scalable} shows the method learning (left column) and subsequently recognising (right column) objects from our experiments.
\begin{figure}[h!]
	\begin{center}
		\includegraphics[width=0.6\columnwidth]{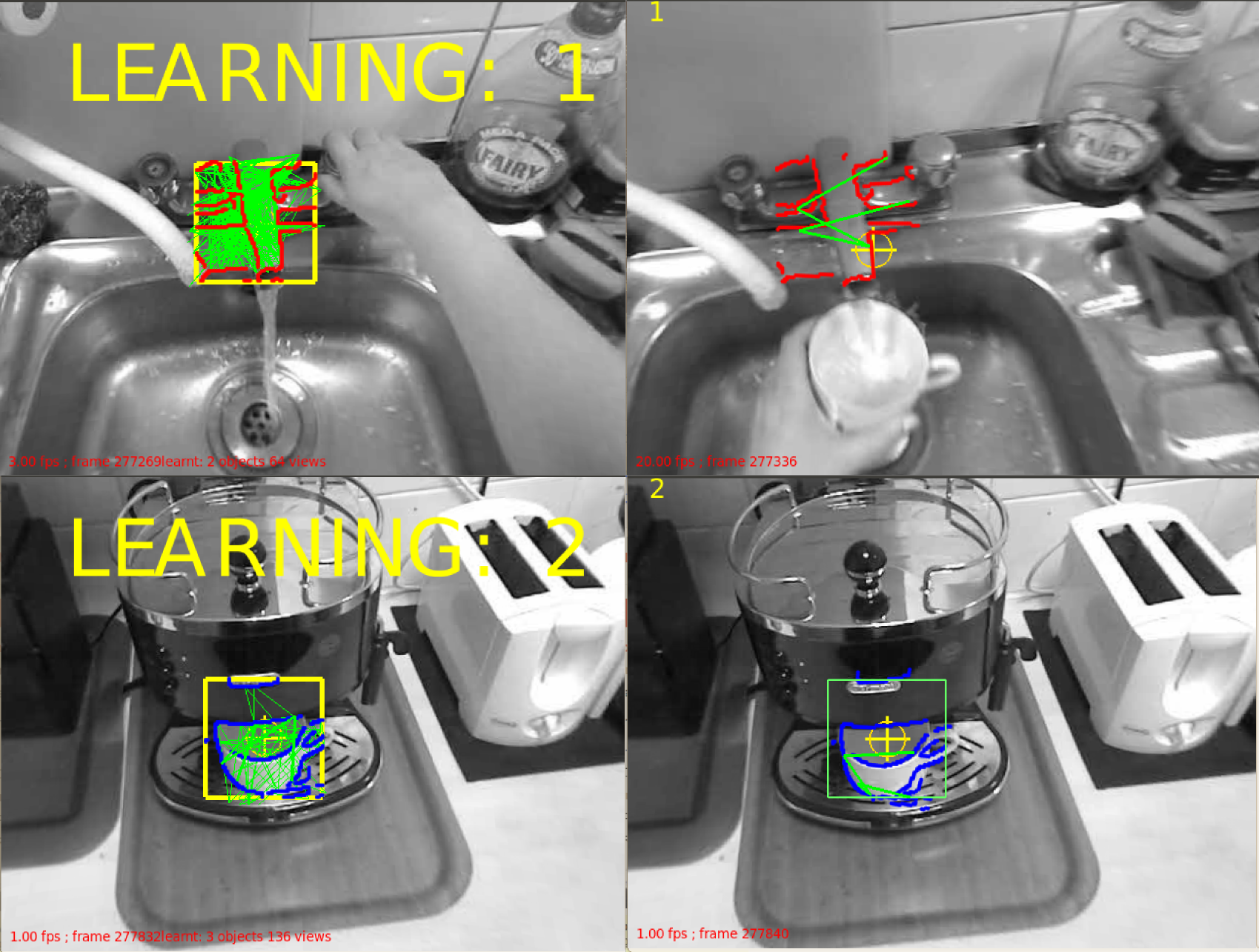}
	\end{center}
	\caption{During discovery (left), edges within $\Omega(I_t)$ are captured as object views, and represented using affine-invariant descriptors~\cite{Damen12}. These are used to detect objects around the gaze point in real-time (right).}
        \label{fig:scalable}
\end{figure}
\end{enumerate}

\section{Experiments and Results}
\label{sec:results}
\begin{figure}[h!]
\centering
\includegraphics[width=0.95\columnwidth]{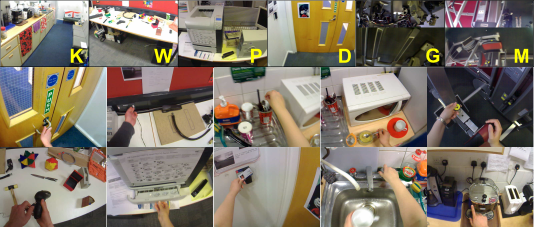}
\caption{Sample images from the Bristol Egocentric Object Interactions Dataset (BEOID). }
\label{fig:beoid}
\end{figure}
\noindent {\textbf {Setup \& Dataset:}} \hspace{4pt}
The wearable gaze tracker hardware (ASL Mobile Eye XG~\cite{ASL}) consists of two cameras sharing a half-mirror, one looking at the scene and another looking at the eye.
After calibration, the scene images are synchronised with, if available, 2D gaze points.
Six locations were chosen: kitchen (K), workspace (W), laser printer (P), corridor with a locked door (D), cardiac gym (G) and weight-lifting machine (M) (Fig.~\ref{fig:beoid}).
For the first four locations (K, W, P, D), sequences from five different operators were recorded, and from three operators for the last two locations (G, M)~\footnote{Dataset available at: \url{http://www.cs.bris.ac.uk/~damen/BEOID}}.
Sample images from the dataset are shown in Fig.~\ref{fig:beoid}.
Following the gaze tracker calibration, the operator moved freely between the locations performing verbally-communicated tasks (Tab.~\ref{tab:dataset}).
Two sequences were recorded for each operator.

\begin{table}[h!]
\begin{center}
{\fontsize{8}{8} \selectfont
\def\arraystretch{1.3}
\begin{tabular}{|l|c|C{30pt}|C{30pt}|C{30pt}|C{30pt}|C{30pt}|C{30pt}|}
\hline
&\textbf{Number of sequences} &\multicolumn{2}{c|}{\textbf{Sequence length}} &\multicolumn{2}{c|}{\textbf{Tracked (\%)}} &\multicolumn{2}{c|}{\textbf{Gaze Fixations (\%)}}\\ \cline{3-8}
& &\textbf{$\mu$} &\textbf{$\sigma$} &\textbf{$\mu$} &\textbf{$\sigma$} &\textbf{$\mu$} &\textbf{$\sigma$} \\ \hline
K &10 &1905 &386 &69.4 &9.1 &58.9 &11.1\\ \cline{2-8}
& \multicolumn{7}{l|}{Prepare coffee using the machine, place the cup on the mat and add sugar}\\
&\multicolumn{7}{l|}{[tap, coffee machine, heat mat, cutlery drainer], (cup, sugar jar)}\\ \hline
W &10 &1221 &194 &78.3 &12.4 &61.9 &18.1\\ \cline{2-8}
&\multicolumn{7}{l|}{Plug the screwdriver for charging and place the tape in the red box [Socket, Box], (screwdriver, charger, tape)} \\ \hline
P &10 &596 &77 &75.8 &13.3 &70.5 &14.1\\ \cline{2-8}
&\multicolumn{7}{l|}{Check the printer is loaded with paper manually and using the keypad [drawer, keypad]}\\ \hline
D &10 &303 &83 &71.8 &15.8 &56.2 &14.7\\ \cline{2-8}
&\multicolumn{7}{l|}{Go through the locked door [door lock, door handle]}\\ \hline
G &6 &5183 &482 &76.4 &9.0 &66.7 &11.0\\ \cline{2-8}
&\multicolumn{7}{l|}{Use the treadmill and the bicycle next to it [treadmill panel, bicycle panel]}\\ \hline
M &6 &2059 &624 &24.5 &16.2 &14.6 &15.2\\ \cline{2-8}
&\multicolumn{7}{l|}{Adjust the seat, chest pad and weight then use the machine [seat adjuster, pad adjuster, weight adjuster]}\\ \hline
\end{tabular}}
\end{center}
\caption{For the six locations, the number of sequences, average number of frames, percentage of tracked frames, percentage of gaze fixations, as well as the verbally communicated tasks, fixed~``[]'' and movable~``()'' ground-truth TROs.}
\label{tab:dataset}
\end{table}
\begin{figure}[h!]
\centering
\includegraphics[width=0.95\columnwidth]{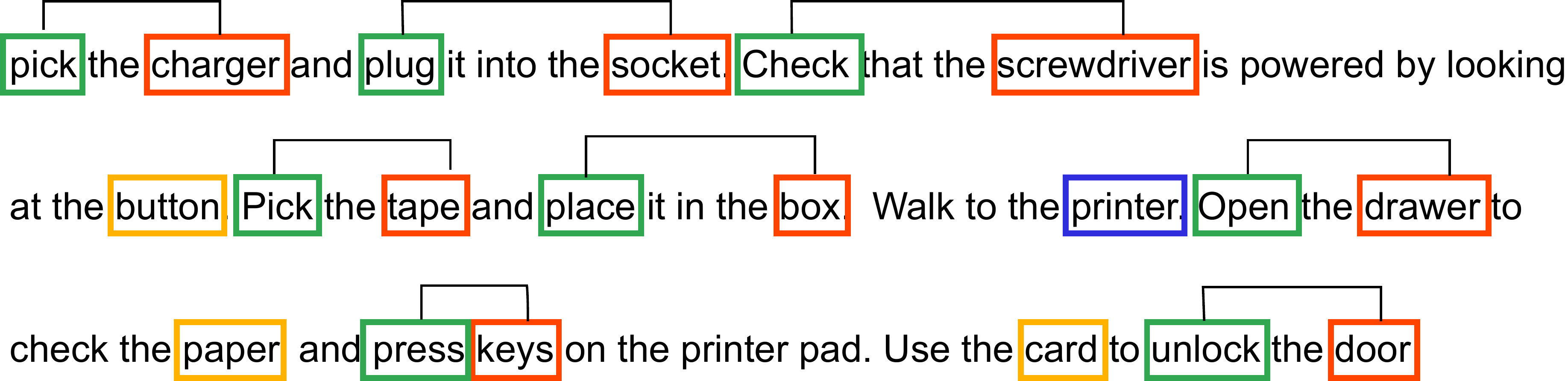}
\caption{Example showing how the ground truth for TROs and MOIs was obtained from subject's narrations. Ground-truth TROs narrated by more than 50\% of subjects are framed in red, compared to less-frequent subjects (orange). Location names are ignored (blue). The verb-noun combinations are used to ground-truth MOIs (green). The narrations are released with the dataset. }
\label{fig:narration}
\end{figure}

The operators were then asked to watch the videos, and write down a narration of what they have performed.
Narrations were stemmed manually to match nouns and verbs which are semantically identical (e.g. adaptor vs. charger, pick vs. retrieve).
Nouns narrated by more than 50\% of the operators represent the twenty ground-truth TROs.
Narrated verb-noun combinations are labelled as MOIs.
Objects varied between having a single MOI (e.g. door handle: open) and up to three different usage methods (e.g. sugar jar: pick, put, get sugar).
Figure~\ref{fig:narration} shows an example of how the narrations were used to generate the ground-truth TROs and MOIs.

For each location, a map of the environment is built using Parallel Tracking and Mapping (PTAM)~\cite{Klein2007}.
A 3D bounding box around each object is manually labelled for evaluation.
For moveable objects, their different locations are ground-truthed.

\noindent {\textbf {Parameters:}} \hspace{4pt}
In all results, the image parts $\Omega(I_t)$ were fixed to a window size of $200 \times 200$ pixels,
This corresponds to $19.3^\circ$ visual angles in the scene camera.
To calculate appearance descriptors, $\Omega(I_t)$ is divided into $10 \times 10$ non-overlapping patches for calculating HOG descriptors.
In offline processing, the number of words in BoW representation is set to 200.
In calculating the BD index, $K = [2 .. 2 No_{gt}]$ (Eq.~\ref{eq:bd_index}) where $No_{gt}$ is the number of ground-truth objects.
In online TRO discovery, $\xi$ was set to 40 frames which corresponds to 1333ms of attention.

\begin{table}[h!]
\begin{center}
{\fontsize{9}{9} \selectfont
\def\arraystretch{1.2}
\begin{tabular}{|l|l|l|l|l||c|c|c|c|c|c|}
\hline
&&\multirow{3}{*}{$w$} &\multirow{3}{*}{clustering} &&\multicolumn{3}{c}{Without Attention $\Omega_c$} &\multicolumn{3}{|c|}{With Attention $\Omega_g$}\\ \cline{6-11}
&& & &&app &pos &both &app &pos &both \\ \hline
\multirow{24}{*}{\rotatebox[origin=c]{90}{\textbf{Offline}}} &\multirow{12}{*}{\rotatebox[origin=c]{90}{Known K}} &\multirow{6}{*}{1} &\multirow{3}{*}{k-means} &Recall &50.5 &55.0 &60.0 &55.0 &80.0 &80.0 \\ \cline{5-11}
& & & &Precision &52.6 &61.1 &66.7 &61.1 &84.2 &84.2 \\ \hhline{*{4}~*{7}-}
& & & &\cellcolor{LightCyan}F-1 Score &\cellcolor{LightCyan}51.5 &\cellcolor{LightCyan}57.9 &\cellcolor{LightCyan}63.2 &\cellcolor{LightCyan}57.9 &\cellcolor{LightCyan}\textbf{82.0} &\cellcolor{LightCyan}{\textbf{82.0}} \\ \hhline{*{3}~*{8}-}
& & &\multirow{3}{*}{Spectral} &Recall &45.0 &60.0 &50.0 &60.0 &80.0 &90.0\\ \cline{5-11}
& & & &Precision &47.4 &66.7 &58.8 &60.0 &80.8 &90.0 \\ \hhline{*{4}~*{7}-}
& & & &\cellcolor{LightCyan}F-1 Score &\cellcolor{LightCyan}46.2 &\cellcolor{LightCyan}63.2 &\cellcolor{LightCyan}54.0 &\cellcolor{LightCyan}60.0 &\cellcolor{LightCyan}80.4 &\cellcolor{LightCyan}\textbf{90.0} \\ \hhline{*{2}~*{9}-}
& &\multirow{6}{*}{25} &\multirow{3}{*}{k-means} &Recall &50.0 &60.0 &55.0 &60.0 &85.0 &85.0\\ \cline{5-11}
& & & &Precision &52.6 &70.6 &64.7 &60.0 &89.5 &89.5 \\ \hhline{*{4}~*{7}-}
& & & &\cellcolor{LightCyan}F-1 Score &\cellcolor{LightCyan}51.3 &\cellcolor{LightCyan}64.9 &\cellcolor{LightCyan}59.5 &\cellcolor{LightCyan}60.0 &\cellcolor{LightCyan}\textbf{87.2} &\cellcolor{LightCyan}\textbf{87.2} \\ \hhline{*{3}~*{8}-}
& & &\multirow{3}{*}{Spectral} &Recall &50.0 &60.0 &55.0 &70.0 &90.0 &\
90.0\\ \cline{5-11}
& & & &Precision &55.6 &66.7 &57.9 &73.7 &90.0 &94.7 \\ \hhline{*{4}~*{7}-}
& & & &\cellcolor{LightCyan}F-1 Score &\cellcolor{LightCyan}52.7 &\cellcolor{LightCyan}63.2 &\cellcolor{LightCyan}56.4 &\cellcolor{LightCyan}71.8 &\cellcolor{LightCyan}90.0 &\cellcolor{LightCyan}\textbf{92.3} \\ \hhline{*{1}~*{10}-}
&\multirow{12}{*}{\rotatebox[origin=c]{90}{DB Index}} &\multirow{6}{*}{1} &\multirow{3}{*}{k-means} &Recall &35.0 &40.0 &40.0 &55.0 &65.0 &65.0\\ \cline{5-11}
& & & &Precision &50.0 &40.0 &44.4 &40.7 &59.1 &61.9 \\ \hhline{*{4}~*{7}-}
& & & &\cellcolor{LightCyan}F-1 Score &\cellcolor{LightCyan}41.2 &\cellcolor{LightCyan}40.0 &\cellcolor{LightCyan}42.1 &\cellcolor{LightCyan}46.8 &\cellcolor{LightCyan}61.9 &\cellcolor{LightCyan}\textbf{63.4} \\ \hhline{*{3}~*{8}-}
& & &\multirow{2}{*}{Spectral} &Recall &50.0 &65.0 &60.0 &65.0 &85.0 &90.0\\ \cline{5-11}
& & & &Precision &41.7 &54.2 &52.2 &41.9 &68.0 &75.0\\ \hhline{*{4}~*{7}-}
& & & &\cellcolor{LightCyan}F-1 Score &\cellcolor{LightCyan}45.5 &\cellcolor{LightCyan}59.1 &\cellcolor{LightCyan}55.8 &\cellcolor{LightCyan}51.0 &\cellcolor{LightCyan}75.6 &\cellcolor{LightCyan}\textbf{81.8} \\ \hhline{*{2}~*{9}-}
& &\multirow{6}{*}{25} &\multirow{3}{*}{k-means} &Recall &60.0 &40.0 &45.0 &60.0 &65.0 &70.0 \\ \cline{5-11}
& & & &Precision &44.4 &42.1 &52.9 &42.9 &59.1  &63.6 \\ \hhline{*{4}~*{7}-}
& & & &\cellcolor{LightCyan}F-1 Score &\cellcolor{LightCyan}51.0 &\cellcolor{LightCyan}41.0 &\cellcolor{LightCyan}48.6 &\cellcolor{LightCyan}50.0 &\cellcolor{LightCyan}61.9 &\cellcolor{LightCyan}\textbf{66.7} \\ \hhline{*{3}~*{8}-}
& & &\multirow{3}{*}{Spectral} &Recall &70.0 &75.0 &60.0 &70.0 &80.0 &95.0\\ \cline{5-11}
& & & &Precision &45.2 &51.7 &50.0 &48.3 &59.3 &73.0 \\ \hhline{*{4}~*{7}-}
& & & &\cellcolor{LightCyan}F-1 Score &\cellcolor{LightCyan}54.9 &\cellcolor{LightCyan}61.2 &\cellcolor{LightCyan}54.5 &\cellcolor{LightCyan}57.2 &\cellcolor{LightCyan}68.1 &\cellcolor{LightCyan}\textbf{82.6} \\ \hline

\multicolumn{4}{|l|}{} &Recall &26.7 &7.2 &40.0 &73.3 &13.3 &85.0 \\\cline{5-11}
\multicolumn{4}{|l|}{\multirow{1}{*}{\textbf{Online}}}&Precision &50.0 &6.7 &52.9 &55.0 &7.2 &77.3\\ \hhline{*{4}~*{7}-}
\multicolumn{4}{|l|}{} &\cellcolor{LightCyan}F-1 Score &\cellcolor{LightCyan}34.8 &\cellcolor{LightCyan}6.9 &\cellcolor{LightCyan}45.6 &\cellcolor{LightCyan}62.8 &\cellcolor{LightCyan}9.3 &\cellcolor{LightCyan}\textbf{81.0} \\ \hline
\end{tabular}
}
\end{center}
\vspace*{-5pt}
\caption{\red{Recall, precision and F1-score} results for discovering TROs using different features, clustering methods, with/without attention and sliding window for the proposed offline and online TRO discovery methods.}
\label{tab:stats}
\end{table}

\begin{figure}[h!]
\centering
\includegraphics[width=0.8\columnwidth]{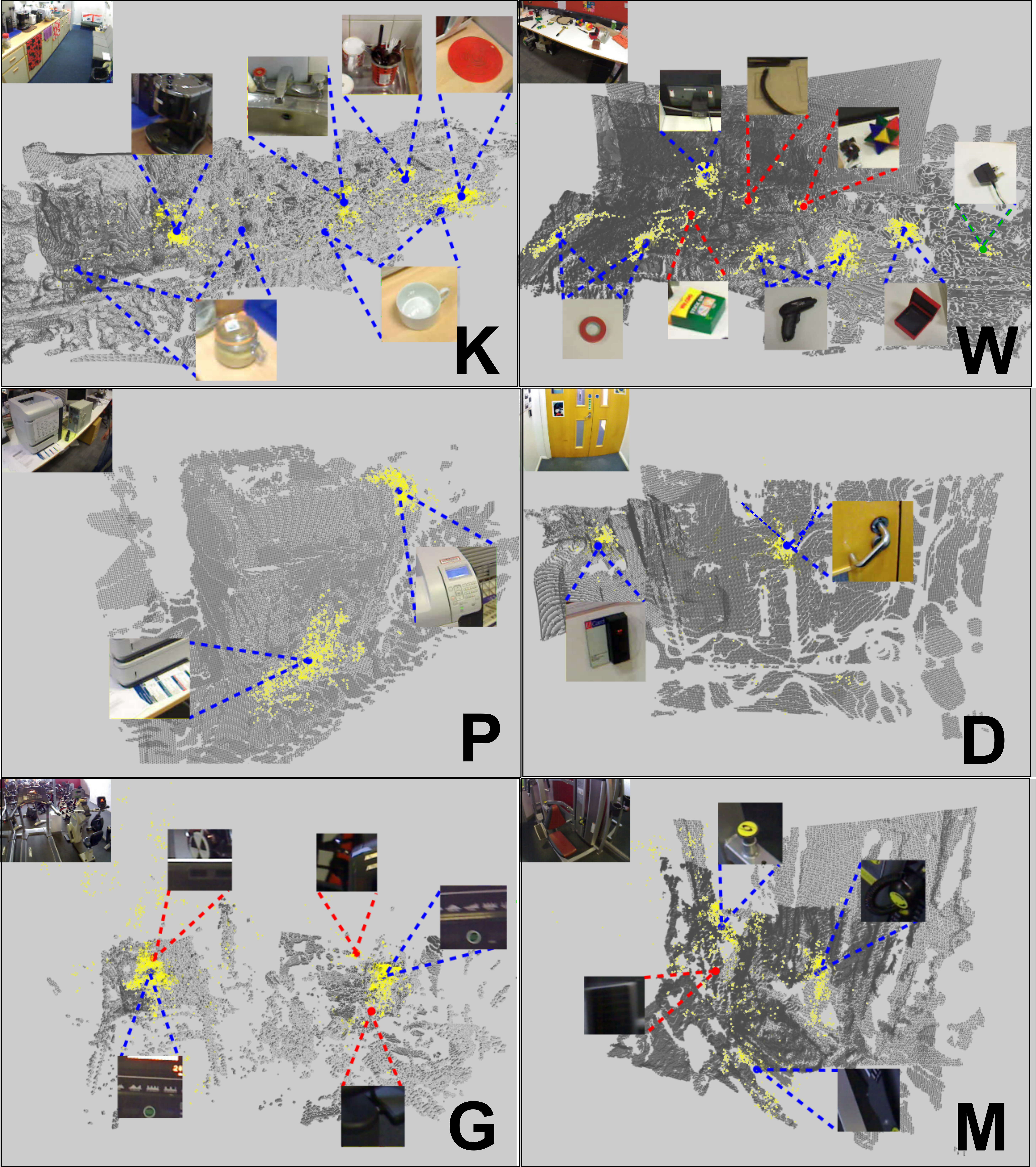}
\caption{Discovered TROs (offline - appearance, position, attention, spectral clustering, $w=25$ and DB index (i.e. number of objects is unknown)). An overview of the locations is shown at the top. Blue dots represent true-positive (19 objs),  red dots represent false positive (7 objs) and green dots represent false negative (1 obj).}
\label{fig:discovered}
\end{figure}

\begin{figure}[h!]
\centering
\includegraphics[width=1.0\columnwidth]{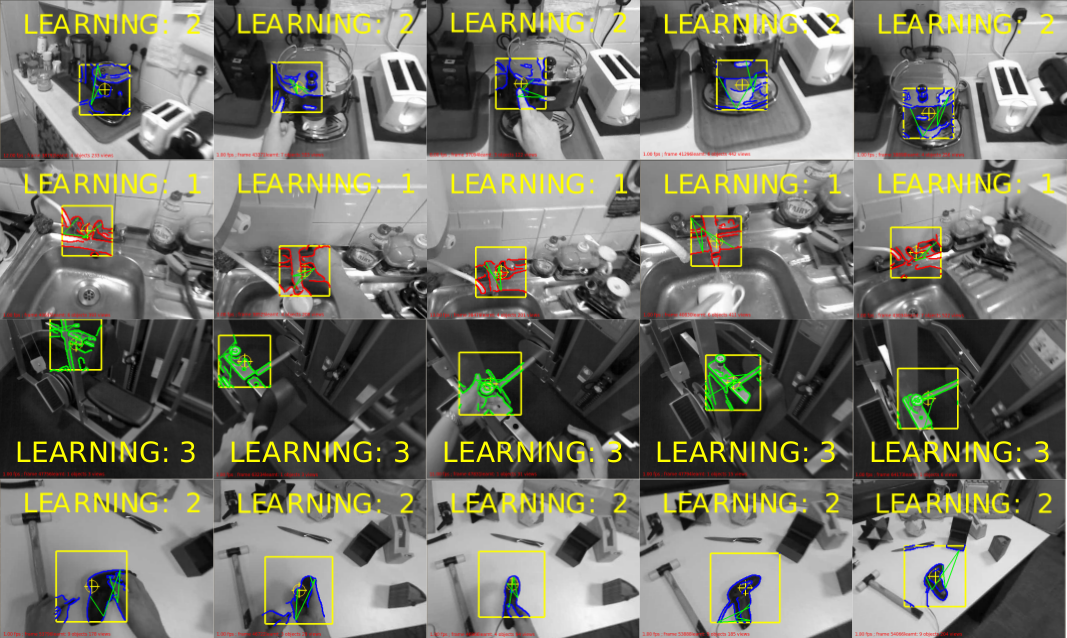}
\caption{Learnt views from training sequences of multiple users for a variety of objects: coffee machine, tap, seat adjuster and screwdriver.}
\label{fig:viewpoints}
\end{figure}

\begin{figure}[h!]
\centering
\includegraphics[width=0.95\columnwidth]{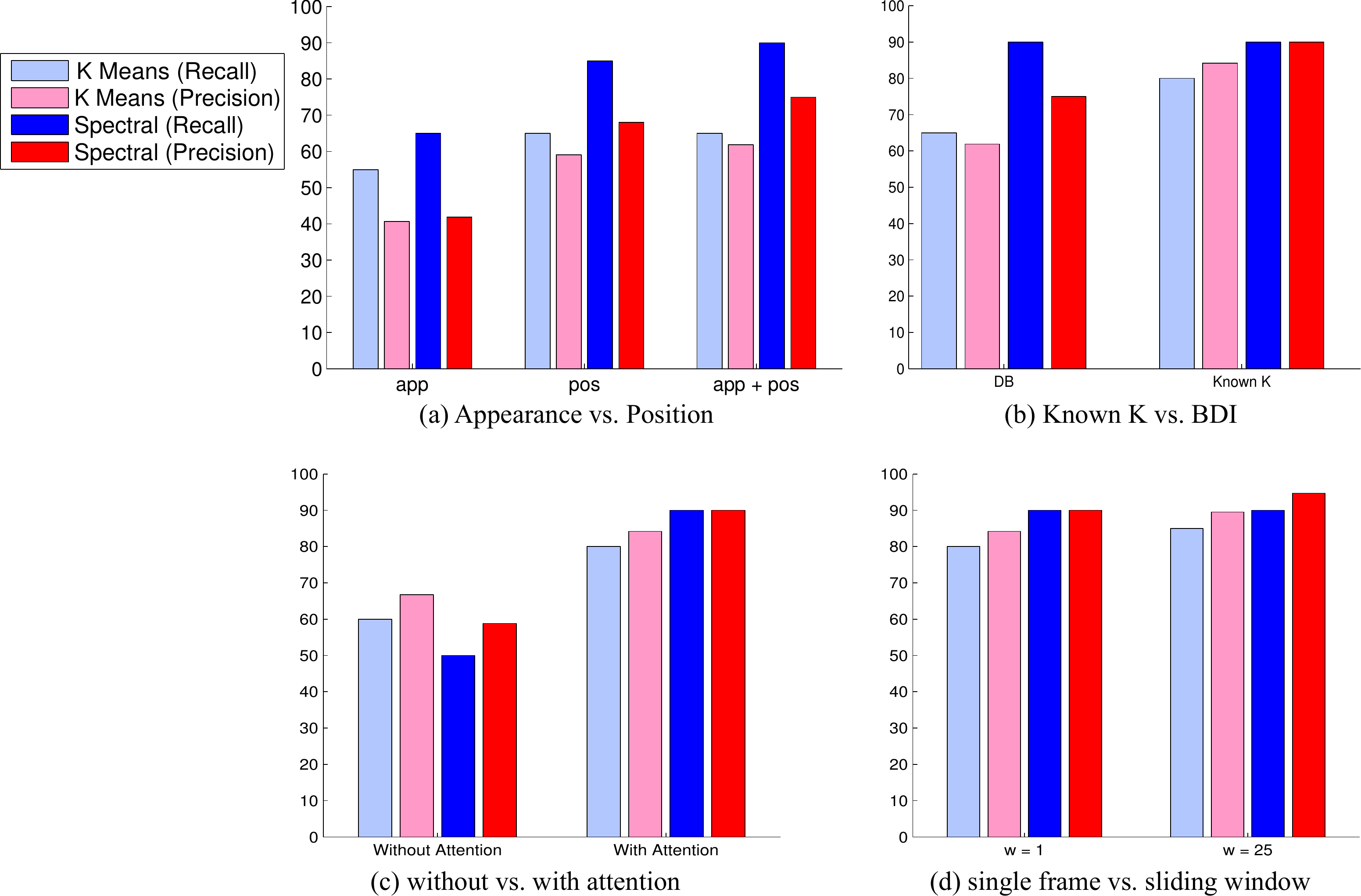}
\caption{\textbf{(a)} appearance (app) vs position (pos) and their combination (app+pos) using spectral vs. k-means clustering using DB index. \textbf{(b)} Using app+pos, DB index vs. known number of clusters. \textbf{(c)} For app+pos+knownK, parts centred around centre of image vs. gaze fixations. \textbf{(d)} Single-frame vs. sliding window representations.}
\label{fig:stats}
\end{figure}

\noindent{\textbf{Results for discovering TROs:}} \hspace{4pt} The results of offline and online TRO discovery are compared to the established ground-truth.
The clusters' 3D bounding boxes are compared to ground-truth bounding boxes and the PASCAL overlap criteria (in 3D) of 20\% indicates a true-positive.
This is because the viewed positions don't typically cover the full extent of the object.
Table~\ref{tab:stats} shows the complete set of results.
In offline TRO discovery, two clustering methods are compared - spectral clustering and k-means.
Appearance and position features are used individually or combined, either for a single frame ($w = 1$) or a sliding window ($w = 25$).
The image part mechanisms $\Omega_c$ and $\Omega_g$ are compared, where the latter crops an image around gaze fixations thus referred to as cropping `with attention'.
Estimating the number of clusters using the Davies-Bouldin (DB) index is compared to knowing the number of clusters \textit{a priori} (ref. \textit{Known K}).
For online results, the best precision for the highest recall is reported as the parameters $(\epsilon_1, \epsilon_2, \epsilon_3)$ are varied.

Table~\ref{tab:stats} shows that the best offline results are obtained using spectral clustering, combining appearance and position, with attention and over a sliding window.
Using Davies-Bouldin (DB) index, 95\% of the TROs were retrieved with 73\% precision.
These discovered TROs are shown in Fig.~\ref{fig:discovered}.
If the number of clusters was known \textit{a priori}, 90\% of TROs would be discovered with 94\% precision.
This is because the optimal number of clusters using DB index was higher than ground-truth $K$, resulting in one more correct object and several false positive clusters.
In online TRO discovery, attention significantly improves the results as the chance of $\xi$ consecutive similar image parts increases.
Interestingly, when combining appearance and position, 85\% of the objects were retrieved with 77\% precision \red{(F-1 score = 81\%)} showing the potentials of the scalable algorithm.
Examples of learnt views for the discovered objects can be found in Fig~\ref{fig:viewpoints}.

Fig.~\ref{fig:stats} highlights several conclusions from the results of offline TRO discovery: (a) shows that for [DB, attention, $w=1$] position achieves better than appearance when used solely.
This is because most of the objects in our dataset (15/20) are fixed objects.
As expected, adding appearance information increases the precision as this clusters instances of moveable objects into a single cluster.
Fig.~\ref{fig:stats} (b) shows that DB index achieves the same recall as Known K when using spectral clustering [app+pos, attention, $w=1$].
Precision increases when K is known - i.e. smaller discarded clusters actually do not represent TROs.
Fig.~\ref{fig:stats} (c) shows the importance of within-image attention [app+pos, KnownK, $w=1$]. A significant drop in recall is observed when the information is gathered around the image centre rather than gaze fixations. 
Fig.~\ref{fig:stats} (d) shows that a sliding window gives a slight improvement in performance.


\noindent{\textbf{Results for discovering MOIs}} \hspace{4pt}
For each discovered object, the usage snippets longer than $\xi = 1s$ are used to build a usage model.
On average, 16.6 usage snippets are extracted for each TRO ($\sigma = $ 7.4).
Notice that these snippets are extracted automatically based on the discovered TRO.
The example shown here is from the online discovery of TROs for the object \textit{tap}.
We vary the threshold $\lambda$ to accept $p({MOI}_j)$ (Eq.~\ref{eq:moi}) and plot recall-precision curves.
A cluster is true-positive if its representative snippet matches one ground-truth MOI; a duplicate match for the same ground-truth MOI is a false-positive.

\begin{figure}[h!]
\centering
\includegraphics[width=0.32\textwidth]{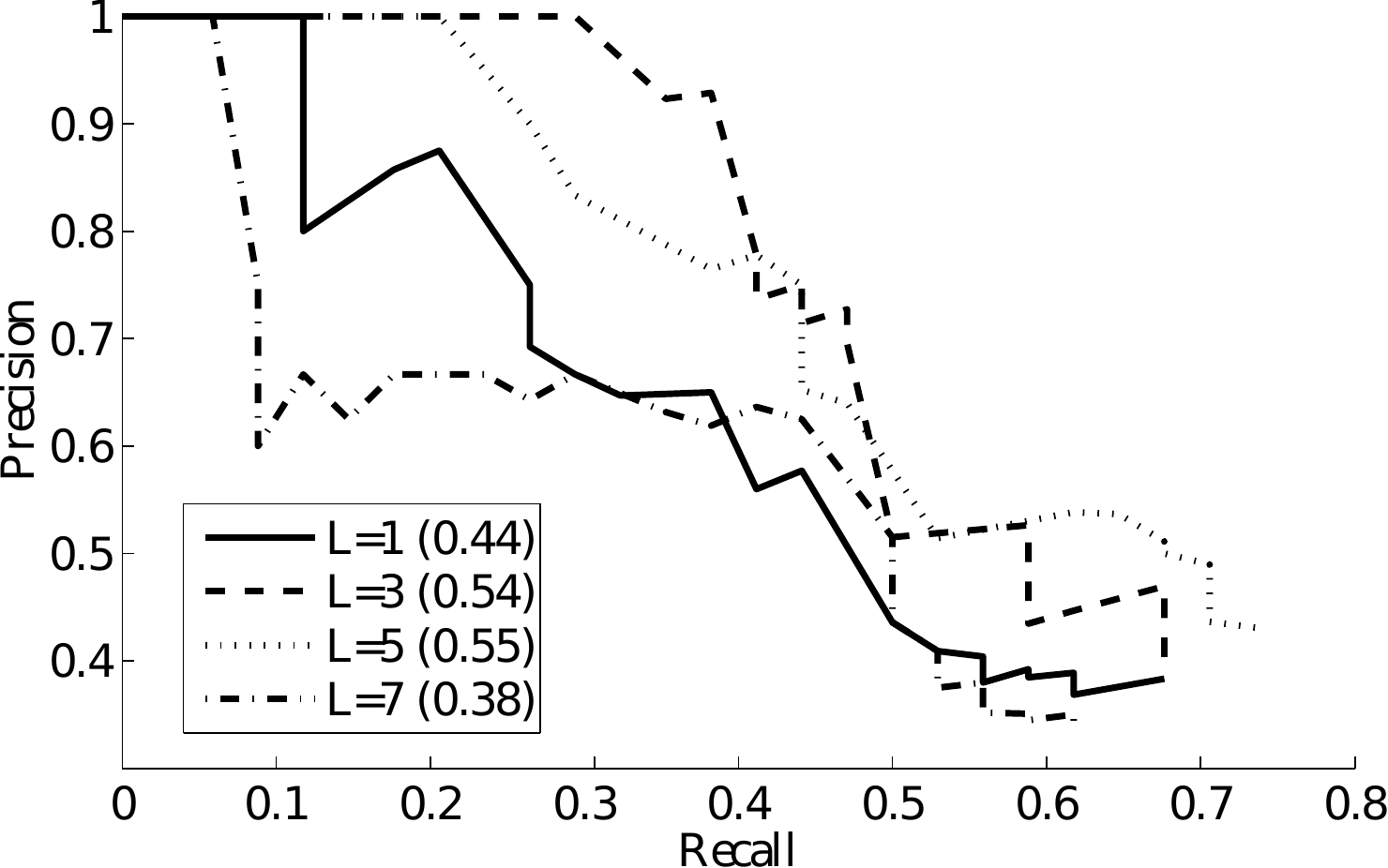}
\includegraphics[width=0.32\textwidth]{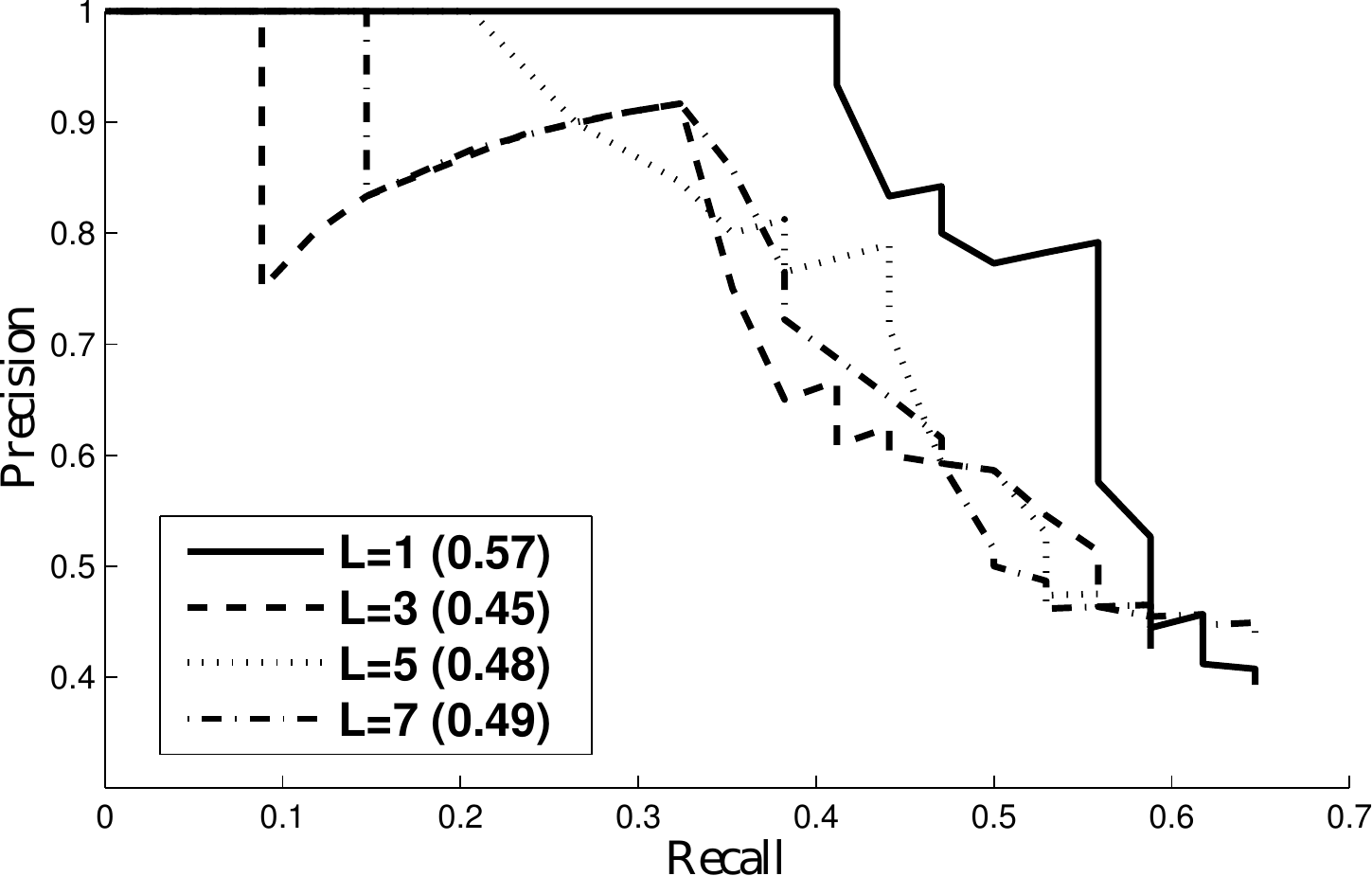}
\includegraphics[width=0.32\textwidth]{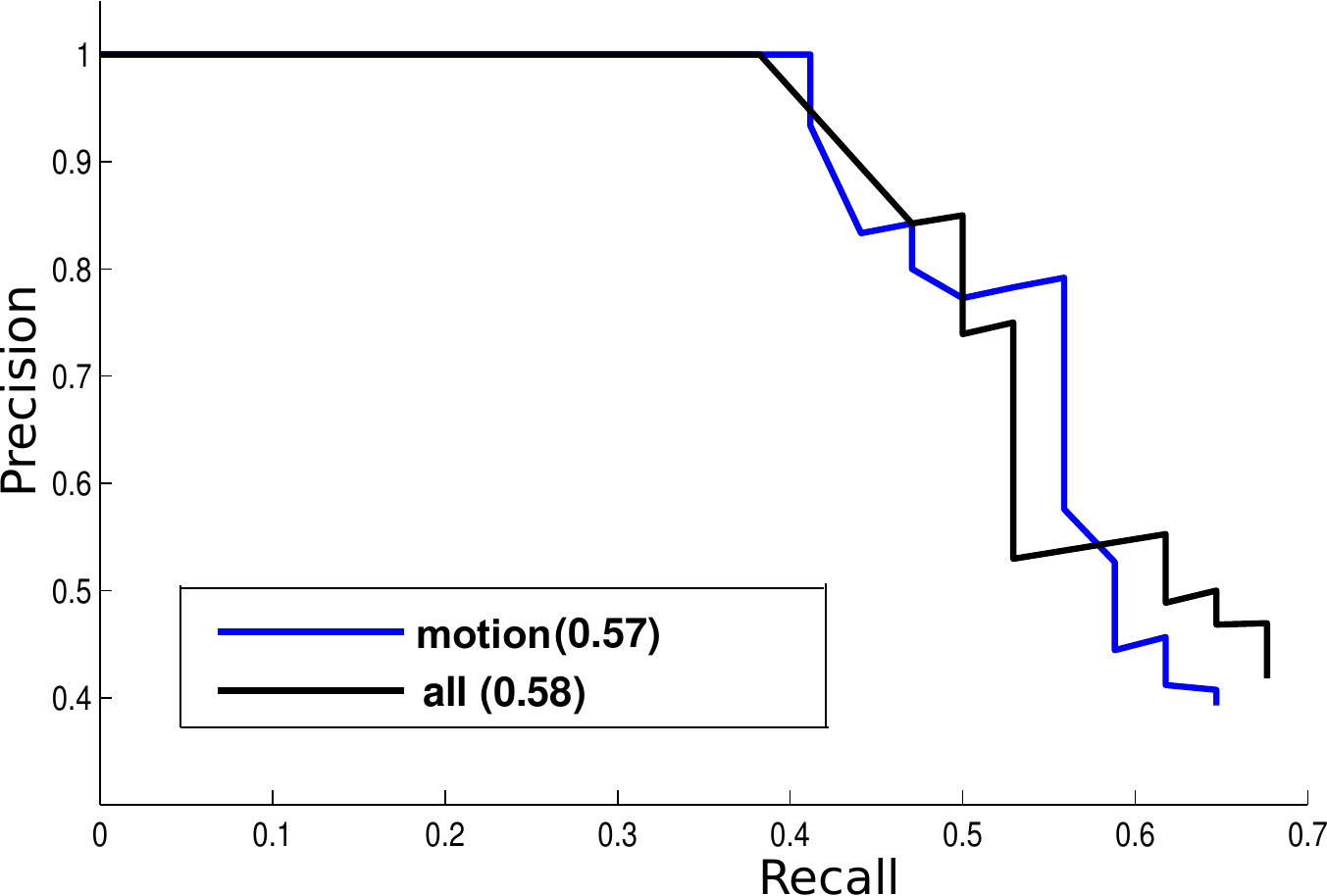}
\caption{For position (left), temporal pyramid (L=5) performed best, while motion (right) performed best on L=1. When using motion only versus combining all features at their best temporal pyramid level, a minor improvement is observed.}
\label{fig:usage_roc_levels}
\end{figure}

\begin{figure}[h!]
\includegraphics[width=1.0\columnwidth]{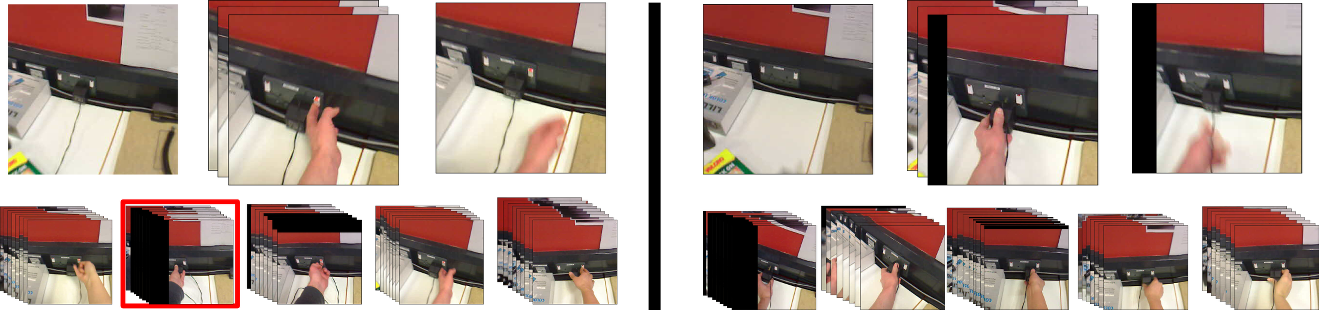}
\caption{For the `socket', the two common MOIs (`switching', `plugging') are found (left \& right). The representative {\it usage snippet} is shown (up) with the other snippets in the same cluster (below) - only one snippet is incorrectly clustered (shown in red).}
\label{fig:usage1}
\end{figure}

\begin{figure}[h!]
\centering
\includegraphics[width=0.85\columnwidth]{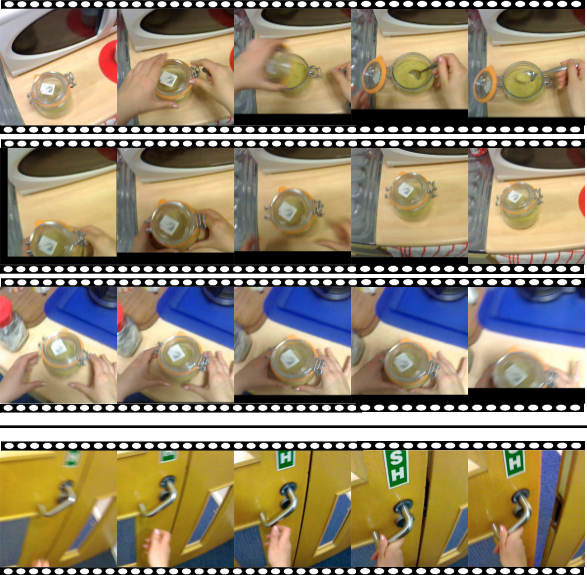}
\caption{For TRO `jar', 3 MOIs are discovered (`get sugar', 'put', 'pick'). For the handle, one MOI is discovered. Frames from the representative snippets are shown.}
\label{fig:usage2}
\end{figure}

We compare using position, appearance and motion features with a temporal pyramid as well as their combination (Fig.~\ref{fig:usage_roc_levels}).
\red{The figure shows that while position information benefits from the temporal pyramid, achieving its highest performance at $L=3$, motion information achieves its best information without using a temporal pyramid $L=1$.
We believe this is due to the various speeds at which people perform the motion.}
As anticipated, motion information solely is capable of distinguishing the various modes of interaction with the same object.
Using the combination of features and $\lambda = 0.2$ \red{(Eq.~\ref{eq:moi})}, the approach is able to discover meaningful MOIs.
Figure~\ref{fig:usage1} shows an example of the method successfully discovering two MOIs for the `socket'.
\red{Given 10 automatically extracted usage snippets, snippets representing the `switch' and `plug' MOIs are separated, with a single snippet incorrectly clustered.
Notice that the motion descriptors are used for clustering without any discriminative tuning to achieve this separation.}
Similarly, Fig.~\ref{fig:usage2} shows further discovered MOIs for the sugar jar and the door handle.
\red{The proposed method is able to discover objects with a single as well as multiple MOIs.
For the sugar jar, the representative usage snippets show the MOIs `get sugar', `put' and `pick' separated.
For the door handle a single MOI is considered common with smaller clusters discarded as spurious.}

\begin{figure}[t]
  \centering
    \includegraphics[width=0.85\columnwidth]{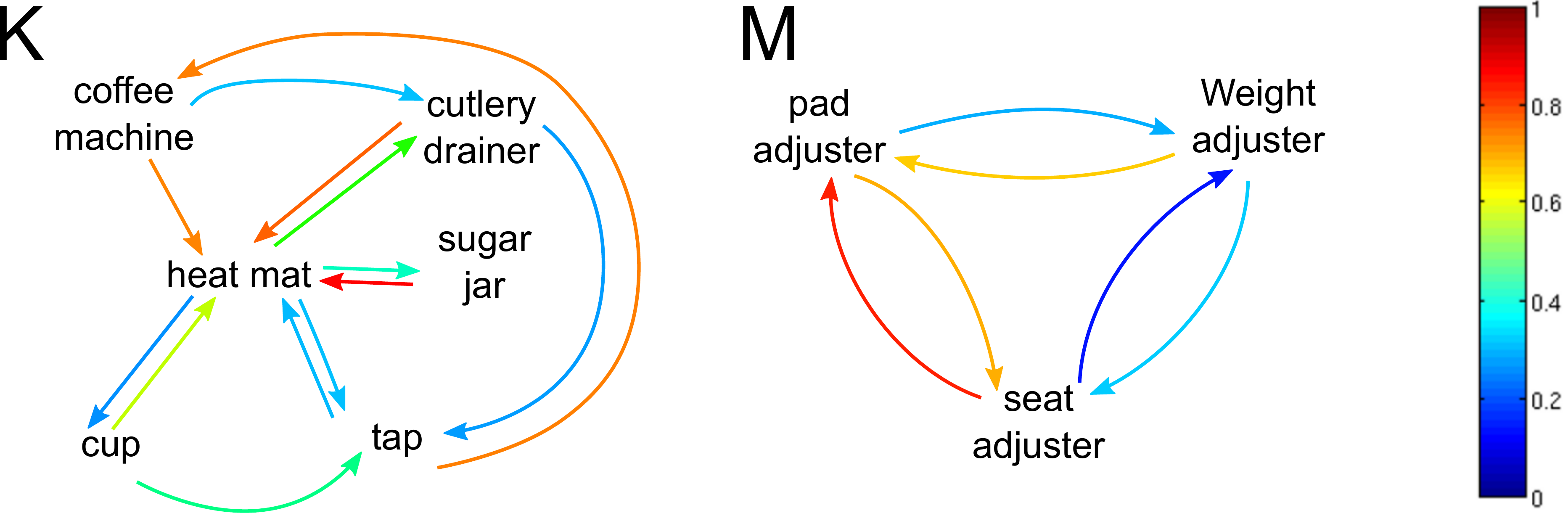}
  \caption{Graph of TRO interactions for two locations kitchen (K) and gym machine (M). Weighting scores of edges in the graph are portrayed as a heat colour map, and edges of weight $\le \alpha$ are not shown.}
  \label{fig:Confusion}
\end{figure}

\noindent{\textbf{Results for Graph of Object Interactions:}} \hspace{4pt} With the discovered TROs, we trained the graphs representing the
interactions. The initial link $\alpha$ was set as
0.05. The generated graphs for the Kitchen (K) and weight Machine (M) sequences are presented in Figure~\ref{fig:Confusion}.
Notice the strong causal links: \textit{coffee machine}$\rightarrow$\textit{heat mat}, \textit{tap}$\rightarrow$\textit{coffee machine}, \textit{sugar jar}$\rightarrow$\textit{heat mat}, \textit{seat adjuster}$\rightarrow$\textit{pad adjuster} all being meaningful strong links between interactions with these objects in the dataset. 

\begin{figure}[h!]
	\begin{center}
		\includegraphics[width=0.7\columnwidth]{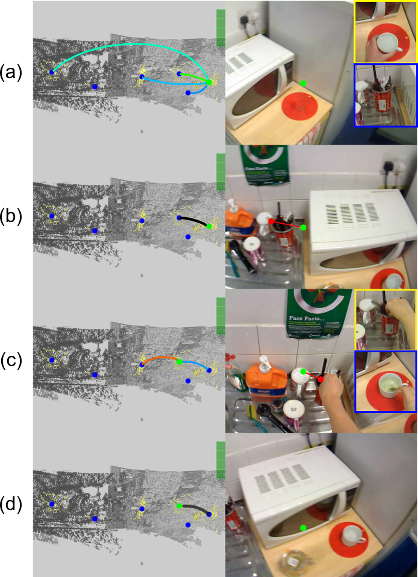}
 	\end{center}
	\caption{For a pre-built map of the environment (left) and egocentric images with tracked gaze (right), the position (green-dot) is used to recognise TRO, a usage snippet is inserted (yellow-framed insert) along with the object to be used next (blue-framed insert). The recommendations are inserted everytime a different TRO is recognised.}
\label{fig:graph-help}
\end{figure}

\noindent{\textbf{Results for Assistive Mode:}} \hspace{4pt} While we do not test the assistive mode with users to evaluate the `usefulness' of the provided usage snippets or the recommendation for the object to use next, we qualitatively assess the ability of the assistive mode to provide meaningful help snippets.
In the assistive mode, we use a leave-one-out; for every operator, TROs are discovered and common MOIs are found from sequences of other operators.
In the assistive mode, when a discovered TRO is detected, an insert is shown indicating a suggestive way of how the object can be used and what object to use next.
We show results from the two recognition methods, first employing the position models to predict the object being used, then employing the appearance models.

Fig~\ref{fig:graph-help} shows a sequence of object interactions in the assistive mode, using location models to recognise TROs.
When the user fixates at a discovered TRO, a usage snippet indicating how to use that object is recommended along with the object to be used next. The figure also shows links (coloured using the heat map in Fig.~\ref{fig:Confusion}) to indicate weight of the edges in the object interactions graph. When the TRO \textit{heat mat} is recognised (Fig~\ref{fig:graph-help}a), a usage snippet is shown recommending a cup to be placed on the mat. The next object to be used is thought to be the TRO \textit{drainer}. The operator indeed moves towards the drainer (Fig~\ref{fig:graph-help}b), and the drainer is recognised (Fig~\ref{fig:graph-help}c). The recommended usage is to pick a cutlery and the suggested next object is the heat mat. The harvested view of the heat mat is that with the cutlery being used. Though this is automatically chosen, it is extracted from the set of usage snippets that follow using the drainer. The attention is indeed shifted to the heat mat (Fig~\ref{fig:graph-help}d).

\begin{figure}[h!]
	\begin{center}
		\includegraphics[width=0.32\columnwidth]{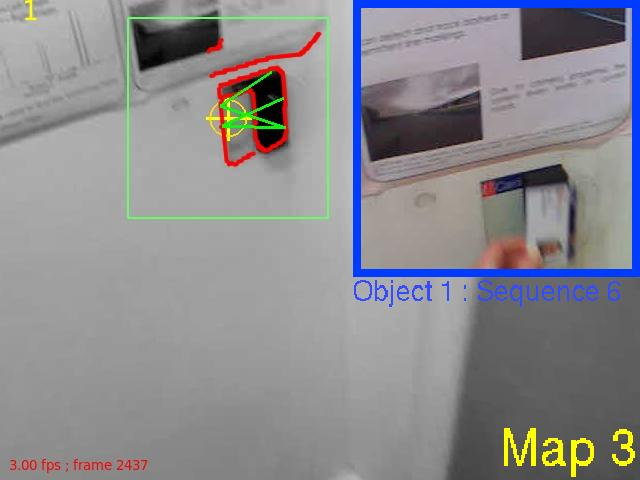}
  	\includegraphics[width=0.32\columnwidth]{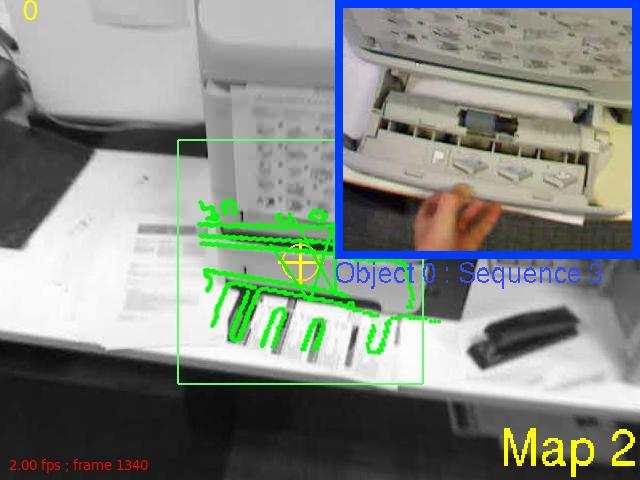}
		\includegraphics[width=0.32\columnwidth]{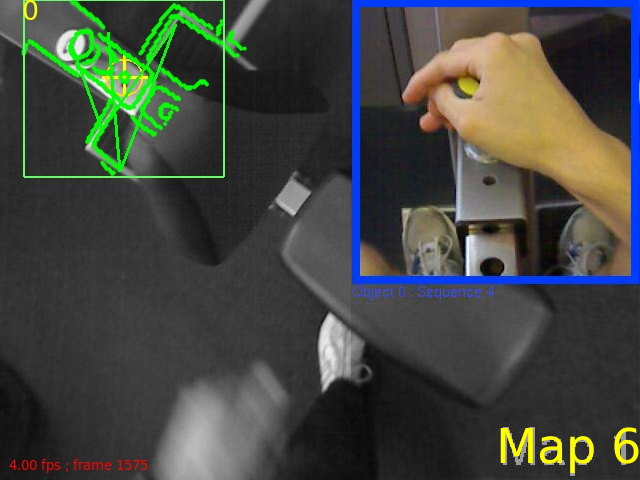}
	\end{center}
	\caption{In the assistive mode, when a TRO is recognised using, usage snippet is inserted (blue-frame) showing the most relevant common MOI based on the initial appearance.}
\label{fig:help}
\end{figure}

Next, we use the real-time texture-minimal scalable detector from~\cite{Damen12} to recognise TROs, due to its light-weight computational load that makes it amendable to wearable systems~\cite{Bunnun12, Hodan15}.
Note that when using the appearance models, a map of the environment would not be needed in the assistive mode.
A {\it help snippet} is displayed each time a new object is recognised.
We showcase video help guides using inserts on a pre-recorded video.
These could in principle be shown on a head-mounted display, but is not considered in this study. 
Figure~\ref{fig:help} shows frames from the help videos and a full sequence is available~\footnote{\url{https://youtu.be/vUeRJmwm7DA}}.
Recall that these inserts are {\it extracted, selected and displayed} fully automatically.
This assistive mode presents a possible application for unsupervised discovery of TROs and their MOIs.
We believe other potential applications could be explored.

\section{Building 3D Models of Task-Relevant Objects}
\label{sec:models}
To build a three dimensional representation of the object $O_k$, we adapt the work of~\cite{Leelasawassuk13} so it does not require the detection of keyframes and it uses input from multiple users.
Given a sparse map of the environment, the 3D points-of-regard are found by back-projecting the rays connecting the camera to the image part.
These are used as seeds for super-pixel segmentation.
The method uses outlier removal to reduce the error in volume estimation.
In this work, we exploit 3D position information to generate textured three-dimensional models of the TROs.
Despite not being perfect models, due to the fact that they are created during task performance, the resulting models are useful visualisations of what objects the system has discovered.
Ultimately, having a 3D model could facilitate applications such as augmented reality guidance.


\begin{figure}[h!]
  \centering
    \includegraphics[width=1.0\columnwidth]{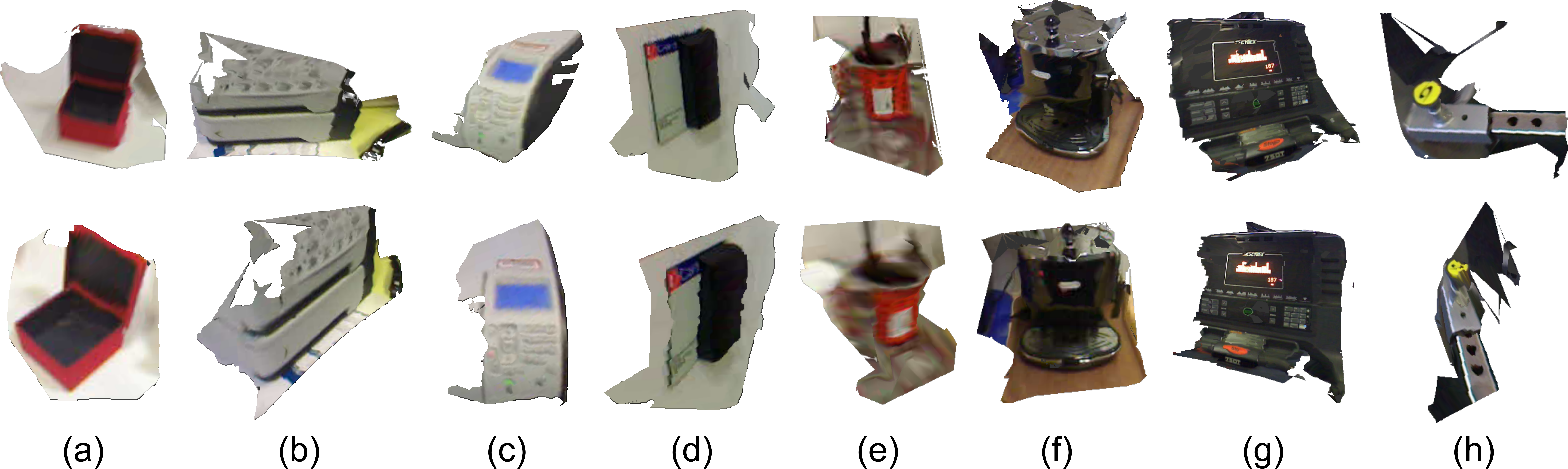}
  \caption{Textured three-dimensional models (two views each) for eight discovered TROs.}
  \label{fig:model3d}
\end{figure}

The accuracy of the model relies on whether it has been viewed from multiple views by the users.
The importance of sequences from multiple users is particularly noticed when attempting to build these approximate 3D models.
Figure~\ref{fig:model3d} presents three-dimensional models for eight discovered objects. 
Note that the method is capable of discovering and representing small-sized (a,d,e,h) as well as larger objects (b,f,g).

\section{Conclusion and Future Work}
In this work, we present an approach for discovering task relevant objects and their common modes of interaction from multi-user egocentric video, {\it fully automatically}.
We compare appearance, position and motion features, along with gaze fixations to indicate attention, for the discovery.
For an unknown number of objects, the approach relies on clustering along with a clustering evaluation measure.
We compare offline clustering to an online algorithm that iteratively refines and updates the clusters.
Both approaches are able to discover fixed objects, such as the sink or a socket as well as moveable objects such as a cup or a sugar jar, as the approaches combine location and appearance features.

On a newly introduced and published egocentric dataset that spans six locations, detailed results show that the offline approach achieved highest performance \red{(F-1 score of 92.3\%)} using spectral clustering over a sliding window, by combining appearance and position features with a gaze fixation attention model.
The online approach achieves \red{an F-1 score of 81\%}. 
Moreover, for each discovered object, various usage snippets have been automatically extracted and clustered using motion features to discover modes of interaction.
First-order Markovian assumption is also employed to build a graph of possible interactions, based on sequences of interactions performed by multiple operators.

Discovered task-relevant objects can be used for providing assistance to users.
As opposed to approaches that require manual authoring of assistance for an object or a task, the assistance proposed here is unsupervised.
Triggered by gazing at the object to be used, the appearance model would recognise a previously discovered task relevant object.
Video-based snippet guidance can then suggest a mode of interaction, given the current status of the object.
This is particularly important for objects that change appearance resulting in varying functionality.
Moreover, a graph of object interactions can be employed to suggest an object to be used next.
This assistance is useful for objects that are often used in consecutive order (e.g. the sink and the drainer or the door lock and the door handle).

The paper provides detailed comparative evaluation for baseline appearance and motion features. State-of-the-art motion (e.g. dense trajectories) and appearance (e.g. convolutional neural networks for tuned discriminative features) features could be investigated. The approach fails to discover objects with very short gaze fixation durations. This is particularly true for objects the user picks up as soon as they are observed. The scalability of the method to discover modes of interaction from multiple tasks requires further research. Currently objects with up to 3 common modes of interaction have been tested.

\red{The paper highlights the importance of attention for discovering task-relevant objects. In this work, we use gaze fixations as a mechanism for detecting attention, both temporally and spatially. Currently, only a few wearable setups offer wearable gaze tracking. Approaches that estimate attention could alternatively be deployed.
Our recent work~\cite{Leelasawassuk15} has detailed a method to estimate attention, both temporally and spatially, from Inertial Unit Measurements (IMU). Alternatively, a method to estimate a visual attention map from the visual flow in an image has been proposed by Matsuo et al~\cite{Matsuo15}. Testing the ability of estimated attention to discover task relevant objects and their modes of interaction is one future direction.}

While this paper provides a complete framework that bridges the gap between unsupervised object discovery and video-based guidance with promising preliminary results, it aims to initiate further research and discussions, particularly related to the usefulness of automatically extracted video guides for human operators and/or autonomous systems, the importance of attention information in egocentric video analysis and more advanced techniques towards discovering modes of interactions for everyday objects.

\vspace{10pt}

\noindent{\textbf{Acknowledgement}} \hspace{4pt} We would like to thank Pished Bunnun, Osian Haines and Andrew Calway for their input on previous iterations of parts of this work. 

\section*{References}
\bibliographystyle{elsarticle-harv}
\bibliography{references}
\end{document}